\documentclass[pdflatex,sn-nature]{sn-jnl}% Style for submissions to Nature Portfolio journals
\usepackage{graphicx}%
\usepackage{multirow}%
\usepackage{amsmath,amssymb,amsfonts}%
\usepackage{amsthm}%
\usepackage{mathrsfs}%
\usepackage[title]{appendix}%
\usepackage{xcolor}%
\usepackage{textcomp}%
\usepackage{manyfoot}%
\usepackage{booktabs}%
\usepackage{algorithm}%
\usepackage{algorithmicx}%
\usepackage{algpseudocode}%
\usepackage{listings}%
\usepackage{lscape}
\theoremstyle{thmstyleone}%
\theoremstyle{thmstyletwo}%

\theoremstyle{thmstylethree}%

\begin{document}

\title[Article Title]{The current status of large language models in summarizing radiology report impressions}


\author*[1]{\fnm{Danqing} \sur{Hu}}\email{hudq@zhejianglab.com}
\equalcont{These authors contributed equally to this work.}

\author[2]{\fnm{Shanyuan} \sur{Zhang}}\email{shanyuanz@163.com}
\equalcont{These authors contributed equally to this work.}

\author[3]{\fnm{Qing} \sur{Liu}}\email{liuqingscn541743@163.com}
\equalcont{These authors contributed equally to this work.}

\author*[1]{\fnm{Xiaofeng} \sur{Zhu}}\email{andy.zhu@zhejianglab.com}

\author*[2]{\fnm{Bing} \sur{Liu}}\email{liubing983811735@126.com}

\affil*[1]{\orgname{Zhejiang Lab}, \orgaddress{\city{Hangzhou}, \postcode{311121}, \state{Zhejiang}, \country{China}}}

\affil*[2]{\orgdiv{Key Laboratory of Carcinogenesis and Translational Research (Ministry of Education), Department of Thoracic Surgery II}, \orgname{Peking University Cancer Hospital and Institute}, \orgaddress{\city{Beijing}, \postcode{100142}, \state{Beijing}, \country{China}}}
\affil[3]{\orgdiv{Department of Radiology}, \orgname{Peking University Cancer Hospital and Institute}, \orgaddress{\city{Beijing}, \postcode{100142}, \state{Beijing}, \country{China}}}


\abstract{Large language models (LLMs) like ChatGPT show excellent capabilities in various natural language processing tasks, especially for text generation. The effectiveness of LLMs in summarizing radiology report impressions remains unclear. In this study, we explore the capability of eight LLMs on the radiology report impression summarization. Three types of radiology reports, i.e., CT, PET-CT, and Ultrasound reports, are collected from Peking University Cancer Hospital and Institute. We use the report findings to construct the zero-shot, one-shot, and three-shot prompts with complete example reports to generate the impressions. Besides the automatic quantitative evaluation metrics, we define five human evaluation metrics, i.e., completeness, correctness, conciseness, verisimilitude, and replaceability, to evaluate the semantics of the generated impressions. Two thoracic surgeons (ZSY and LB) and one radiologist (LQ) compare the generated impressions with the reference impressions and score each impression under the five human evaluation metrics. Experimental results show that there is a gap between the generated impressions and reference impressions. Although the LLMs achieve comparable performance in completeness and correctness, the conciseness and verisimilitude scores are not very high. Using few-shot prompts can improve the LLMs' performance in conciseness and verisimilitude, but the clinicians still think the LLMs can not replace the radiologists in summarizing the radiology impressions.}




\maketitle

\section{Introduction}\label{sec1}

Recently, large language models (LLMs) such as ChatGPT \cite{ChatGPT2022} and GPT-4 \cite{gpt42023} have captured worldwide attention due to their astonishing text-generation capabilities. Through pre-training on vast amounts of data, LLMs demonstrate remarkable  performance on unseen downstream tasks using zero-shot, one-shot, or few-shot prompts without parameter updates \cite{Brown2020}. By reinforcement fine-tuning with human feedback (RLHF) \cite{Ouyang2022}, the LLMs are further guaranteed to produce harmless and unbiased content that aligns with human expectations. The great success of prompt-based LLMs has led to a paradigm shift in NLP research \cite{Tang2023,Hu2024,Hu2024Improving,Chen2023large,Doshi2024,Keloth2024}, thereby bringing new opportunities for the radiology report impression summarization.

Radiology reports document key information in patients' imaging data, such as CT scans, PET, MRI, X-rays, and ultrasound examinations. Typically, radiology reports consist of two main parts, i.e., findings and impressions. The findings section describes the radiologist's observations in the images, while the impressions section summarizes these observations and provides corresponding diagnoses. Impression summarization refers to the process of condensing the lengthy and detailed findings into concise and informative impressions \cite{Zhang2018,Jiang2023,Cai2023}, which is one of the most crucial applications of text summarization in the medical field \cite{Tian2024}. 

With the rise of prompt-based LLMs, it is an interesting question to explore their capability of summarizing radiology report impressions using zero-shot or few-shot prompts. Although some studies have applied prompt-based LLMs to this task \cite{Liu2023evaluating,Sun2023,Van2024}, they only focus on limited types of reports, typically the X-ray reports, and lack detailed clinical expert evaluation of the generated results \cite{Liu2023evaluating,Sun2023} or only evaluate the LLMs on English reports in a zero-shot manner \cite{Van2024}.  

In this study, we conduct a systematic study to explore the capability of prompt-based LLMs in summarizing the impressions of various types of Chinese radiology reports using zero-shot and few-shot prompts. By leveraging automatic quantitative and clinical expert evaluations, we aim to clarify the current status of LLMs in Chinese radiology report impression summarization and the gap between the current achievements and requirements for application in clinical practice.

\section{Results}

\begin{figure}[h]
\centering
\includegraphics[width=1.0\textwidth]{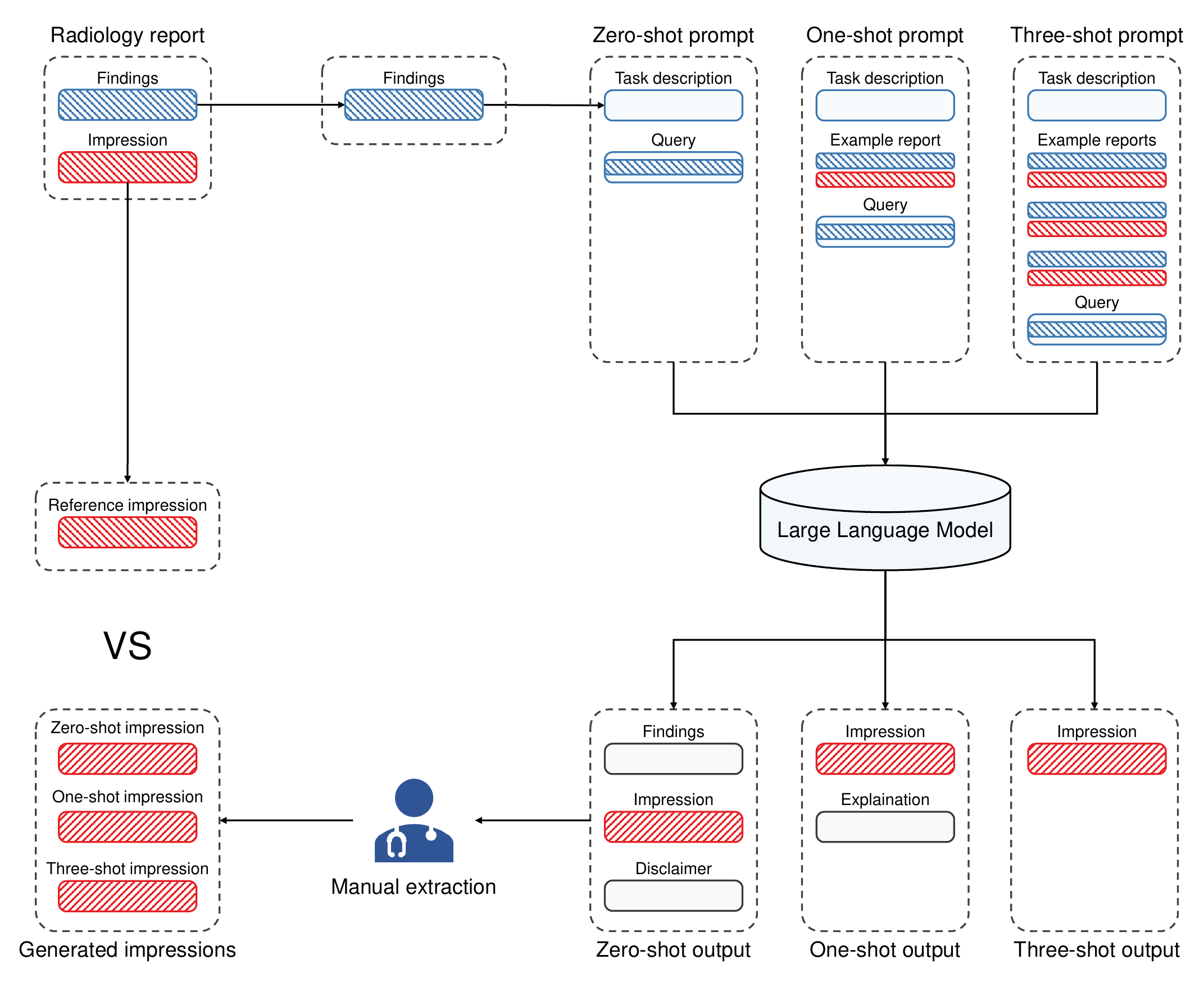}
\caption{The overall pipeline of impression summarization and evaluation.}\label{fig1}
\end{figure}

\subsection{Study overview}

To evaluate the LLMs for impression summarization, we first collect three types of Chinese radiology reports, i.e., PET-CT, CT, and ultrasound reports from Peking University Cancer Hospital and Institute. We randomly sample 100 reports from each type of report as the experimental datasets.

Using the collected reports, we evaluate the zero-shot, one-shot, and three-shot performance of impression summarization of four commercially available LLMs, including Tongyi Qianwen, ERNIE Bot, ChatGPT, Bard, and four open source LLMs, including Baichuan, ChatGLM, HuatuoGPT, and ChatGLM-Med. The zero-shot prompt consists of two parts, i.e., task description and query. We add one example report and three example reports between the task description and query parts as the one-shot and three-shot prompts, respectively. Since the maximun input text lengths supported by LLMs are different, to fairly evaluate and compare the performance of LLMs, we do not conduct experiments when some prompt exceed the maximum input text length of LLMs (one-shot PET-CT prompt for ERNIE Bot, ChatGLM\_Med, three-shot PET-CT prompt for ERINE Bot, Baichuan, HuatuoGPT, and ChatGLM\_Med).

Since the LLMs' output not only contains the generated impression but also contains some content unrelated to the impression, such as the findings, the disclaimer, and the explanation of the response, or repeated text. We manually extract the impression-related content from the outputs for the automatic quantitative and human evaluations. The overall pipeline is shown in Figure \ref{fig1}.

\begin{figure}[h]
\centering
\includegraphics[width=\textwidth]{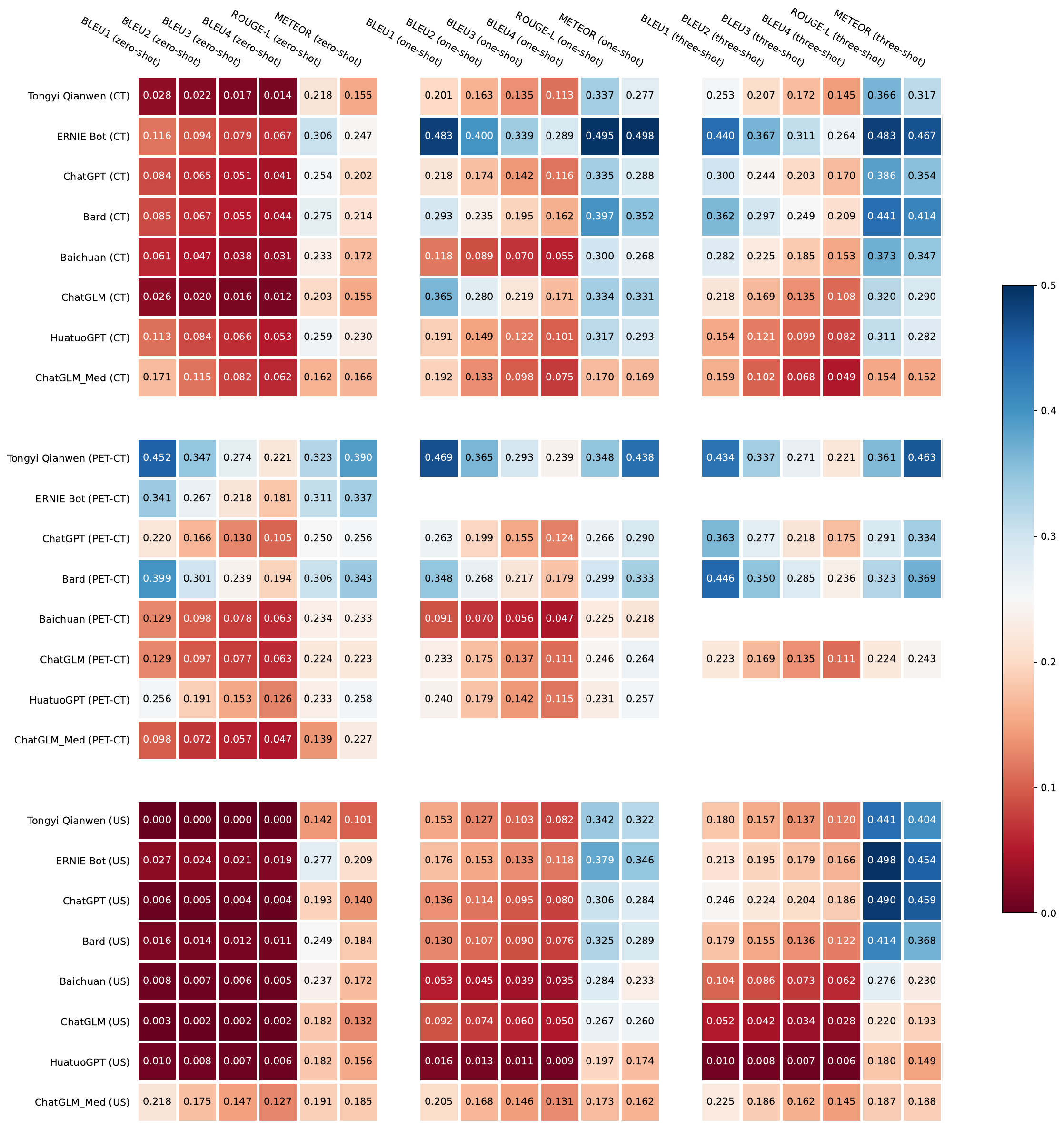}
\caption{The automatic quantitative evaluation results.}\label{fig2}
\end{figure}

\begin{table}[h]
\tiny
\caption{Automatic quantitative evaluation results of the generated CT impressions.}\label{table1}
\begin{tabular*}{\textwidth}{@{\extracolsep\fill}llcccccc}
\toprule%
Prompt type                          & Model              & BLEU1 & BLEU2 & BLEU3 & BLEU4 & ROUGE-L & METEOR  \\
\midrule
\multirow{8}{*}{Zero-shot}           & Tongyi Qianwen     & 0.028 & 0.022 & 0.017 & 0.014 & 0.218 & 0.155\\
                                     & ERNIE Bot          & 0.116 & 0.094 & 0.079 & 0.067 & 0.306 & 0.247\\
                                     & ChatGPT            & 0.084 & 0.065 & 0.051 & 0.041 & 0.254 & 0.202\\
                                     & Bard               & 0.085 & 0.067 & 0.055 & 0.044 & 0.275 & 0.214\\
                                     & Baichuan           & 0.061 & 0.047 & 0.038 & 0.031 & 0.233 & 0.172\\
                                     & ChatGLM            & 0.026 & 0.020 & 0.016 & 0.012 & 0.203 & 0.155\\
                                     & HuatuoGPT          & 0.113 & 0.084 & 0.066 & 0.053 & 0.259 & 0.230\\
                                     & ChatGLM-Med        & 0.171 & 0.115 & 0.082 & 0.062 & 0.162 & 0.166\\
\cmidrule{2-8}
\multirow{8}{*}{One-shot}            & Tongyi Qianwen     & 0.201 & 0.163 & 0.135 & 0.113 & 0.337 & 0.277\\
                                     & \textbf{ERNIE Bot}          & \textbf{0.483} & \textbf{0.400} & \textbf{0.339} & \textbf{0.289} & \textbf{0.495} & \textbf{0.498}\\
                                     & ChatGPT            & 0.218 & 0.174 & 0.142 & 0.116 & 0.335 & 0.288\\
                                     & Bard               & 0.293 & 0.235 & 0.195 & 0.162 & 0.397 & 0.352\\
                                     & Baichuan           & 0.118 & 0.089 & 0.070 & 0.055 & 0.300 & 0.268\\
                                     & ChatGLM            & 0.365 & 0.280 & 0.219 & 0.171 & 0.334 & 0.331\\
                                     & HuatuoGPT          & 0.191 & 0.149 & 0.122 & 0.101 & 0.317 & 0.293\\
                                     & ChatGLM-Med        & 0.192 & 0.133 & 0.098 & 0.075 & 0.170 & 0.169\\
\cmidrule{2-8}
\multirow{8}{*}{Three-shot}          & Tongyi Qianwen     & 0.253 & 0.207 & 0.172 & 0.145 & 0.366 & 0.317\\
                                     & ERNIE Bot          & 0.440 & 0.367 & 0.311 & 0.264 & 0.483 & 0.467\\
                                     & ChatGPT            & 0.300 & 0.244 & 0.203 & 0.170 & 0.386 & 0.354\\
                                     & Bard               & 0.362 & 0.297 & 0.249 & 0.209 & 0.441 & 0.414\\
                                     & Baichuan           & 0.282 & 0.225 & 0.185 & 0.153 & 0.373 & 0.347\\
                                     & ChatGLM            & 0.218 & 0.169 & 0.135 & 0.108 & 0.320 & 0.290\\
                                     & HuatuoGPT          & 0.154 & 0.121 & 0.099 & 0.082 & 0.311 & 0.282\\
                                     & ChatGLM-Med        & 0.159 & 0.102 & 0.068 & 0.049 & 0.154 & 0.152\\
\botrule
\end{tabular*}
\end{table}

\begin{table}[h]
\tiny
\caption{Automatic quantitative evaluation results of the generated PET-CT impressions.}\label{table2}
\begin{tabular*}{\textwidth}{@{\extracolsep\fill}llcccccc}
\toprule%
Prompt type                          & Model              & BLEU1 & BLEU2 & BLEU3 & BLEU4 & ROUGE-L & METEOR  \\
\midrule
\multirow{8}{*}{Zero-shot}           & Tongyi Qianwen     & 0.452 & 0.347 & 0.274 & 0.221 & 0.323 & 0.390\\
                                     & ERNIE Bot          & 0.341 & 0.267 & 0.218 & 0.181 & 0.311 & 0.337\\
                                     & ChatGPT            & 0.220 & 0.166 & 0.130 & 0.105 & 0.250 & 0.256\\
                                     & Bard               & 0.399 & 0.301 & 0.239 & 0.194 & 0.306 & 0.343\\
                                     & Baichuan           & 0.129 & 0.098 & 0.078 & 0.063 & 0.234 & 0.233\\
                                     & ChatGLM            & 0.129 & 0.097 & 0.077 & 0.063 & 0.224 & 0.223\\
                                     & HuatuoGPT          & 0.256 & 0.191 & 0.153 & 0.126 & 0.233 & 0.258\\
                                     & ChatGLM-Med        & 0.098 & 0.072 & 0.057 & 0.047 & 0.139 & 0.227\\
\cmidrule{2-8}
\multirow{8}{*}{One-shot}            & \textbf{Tongyi Qianwen}     & \textbf{0.469} & \textbf{0.365} & \textbf{0.293} & \textbf{0.239} & 0.348 & 0.438\\
                                     & ERNIE Bot          & - & - & - & - & - & -\\
                                     & ChatGPT            & 0.263 & 0.199 & 0.155 & 0.124 & 0.266 & 0.290\\
                                     & Bard               & 0.348 & 0.268 & 0.217 & 0.179 & 0.299 & 0.333\\
                                     & Baichuan           & 0.091 & 0.070 & 0.056 & 0.047 & 0.225 & 0.218\\
                                     & ChatGLM            & 0.233 & 0.175 & 0.137 & 0.111 & 0.246 & 0.264\\
                                     & HuatuoGPT          & 0.240 & 0.179 & 0.142 & 0.115 & 0.231 & 0.257\\                   & ChatGLM-Med        & - & - & - & - & - & -\\  
\cmidrule{2-8}
\multirow{8}{*}{Three-shot}          & \textbf{Tongyi Qianwen}     & 0.434 & 0.337 & 0.271 & 0.221 & \textbf{0.361} & \textbf{0.463}\\
                                     & ERNIE Bot          & - & - & - & - & - & -\\
                                     & ChatGPT            & 0.363 & 0.277 & 0.218 & 0.175 & 0.291 & 0.334\\
                                     & Bard               & 0.446 & 0.350 & 0.285 & 0.236 & 0.323 & 0.369\\
                                     & Baichuan           & - & - & - & - & - & -\\
                                     & ChatGLM            & 0.223 & 0.169 & 0.135 & 0.111 & 0.224 & 0.243\\
                                     & HuatuoGPT          & - & - & - & - & - & -\\
                                     & ChatGLM-Med        & - & - & - & - & - & -\\
\botrule
\end{tabular*}
\end{table}

\begin{table}[h]
\tiny
\caption{Automatic quantitative evaluation results of the generated ultrasound impressions.}\label{table3}
\begin{tabular*}{\textwidth}{@{\extracolsep\fill}llcccccc}
\toprule%
Prompt type                          & Model              & BLEU1 & BLEU2 & BLEU3 & BLEU4 & ROUGE-L & METEOR  \\
\midrule
\multirow{8}{*}{Zero-shot}           & Tongyi Qianwen     & 0.000 & 0.000 & 0.000 & 0.000 & 0.142 & 0.101\\
                                     & ERNIE Bot          & 0.027 & 0.024 & 0.021 & 0.019 & 0.277 & 0.209\\
                                     & ChatGPT            & 0.006 & 0.005 & 0.004 & 0.004 & 0.193 & 0.140\\
                                     & Bard               & 0.016 & 0.014 & 0.012 & 0.011 & 0.249 & 0.184\\
                                     & Baichuan           & 0.008 & 0.007 & 0.006 & 0.005 & 0.237 & 0.172\\
                                     & ChatGLM            & 0.003 & 0.002 & 0.002 & 0.002 & 0.182 & 0.132\\
                                     & HuatuoGPT          & 0.010 & 0.008 & 0.007 & 0.006 & 0.182 & 0.156\\
                                     & ChatGLM-Med        & 0.218 & 0.175 & 0.147 & 0.127 & 0.191 & 0.185\\
\cmidrule{2-8}
\multirow{8}{*}{One-shot}            & Tongyi Qianwen     & 0.153 & 0.127 & 0.103 & 0.082 & 0.342 & 0.322\\
                                     & ERNIE Bot          & 0.176 & 0.153 & 0.133 & 0.118 & 0.379 & 0.346\\
                                     & ChatGPT            & 0.136 & 0.114 & 0.095 & 0.080 & 0.306 & 0.284\\
                                     & Bard               & 0.130 & 0.107 & 0.090 & 0.076 & 0.325 & 0.289\\
                                     & Baichuan           & 0.053 & 0.045 & 0.039 & 0.035 & 0.284 & 0.233\\
                                     & ChatGLM            & 0.092 & 0.074 & 0.060 & 0.050 & 0.267 & 0.260\\
                                     & HuatuoGPT          & 0.016 & 0.013 & 0.011 & 0.009 & 0.197 & 0.174\\
                                     & ChatGLM-Med        & 0.205 & 0.168 & 0.146 & 0.131 & 0.173 & 0.162\\
\cmidrule{2-8}
\multirow{8}{*}{Three-shot}          & Tongyi Qianwen     & 0.180 & 0.157 & 0.137 & 0.120 & 0.441 & 0.404\\
                                     & \textbf{ERNIE Bot}          & 0.213 & 0.195 & 0.179 & 0.166 & \textbf{0.498} & 0.454\\
                                     & \textbf{ChatGPT}            & \textbf{0.246} & \textbf{0.224} & \textbf{0.204} & \textbf{0.186} & 0.490 & \textbf{0.459}\\
                                     & Bard               & 0.179 & 0.155 & 0.136 & 0.122 & 0.414 & 0.368\\
                                     & Baichuan           & 0.104 & 0.086 & 0.073 & 0.062 & 0.276 & 0.230\\
                                     & ChatGLM            & 0.052 & 0.042 & 0.034 & 0.028 & 0.220 & 0.193\\
                                     & HuatuoGPT          & 0.010 & 0.008 & 0.007 & 0.006 & 0.180 & 0.149\\
                                     & ChatGLM-Med        & 0.225 & 0.186 & 0.162 & 0.145 & 0.187 & 0.188\\
\botrule
\end{tabular*}
\end{table}

\subsection{Quantitative evaluation}

To evaluate the generated impressions, we first employ three widely used text summarization evaluation metrics, including BLEU, ROUGE-L, and METEOR, to compare them with the reference impressions. Tables \ref{table1}, \ref{table2}, and \ref{table3} show the experimental results. We notice that ERNIE Bot obtains the overall best results for CT impression summarization, Tongyi Qianwen achieves the best performance for PET-CT impression summarization, and ChatGPT shows the best performance for ultrasound impression summarization. Note that the best LLMs are all commercially available models. Although the Bard model does not obtain the best result for any task, it achieves the second-best results in the PET-CT and CT impression summarization. Figure \ref{fig2} illustrates the experimental results more intuitively. Moreover, all the best results are obtained based on the few-shot prompts, indicating LLMs can learn from the example reports in the prompt to generate better impressions. Figure \ref{fig2} also illustrates that most LLMs can benefit from the few-shot examples in the prompts, but more is not necessarily better.

\subsection{Human evaluation}

Besides automatic quantitative evaluation, in this study, we also conduct a human evaluation to obtain more insights into the impression summarization capabilities of LLMs.

\begin{figure}[h]
\centering
\includegraphics[width=1\textwidth]{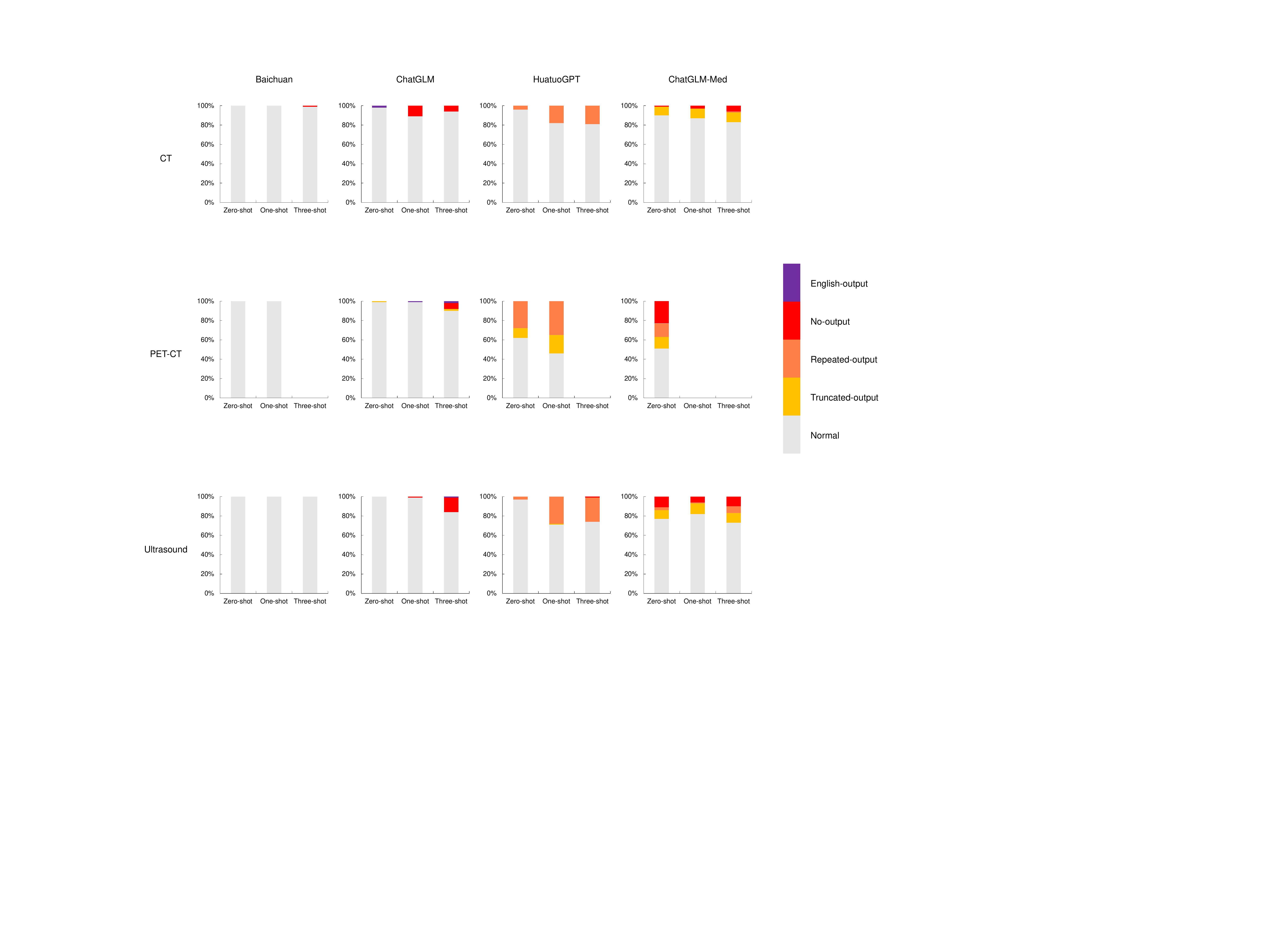}
\caption{Quality evaluation results of the generated impressions.}\label{fig3}
\end{figure}

\subsubsection{Quality evaluation}

As the LLMs may produce undesired context, we first manually review the quality of the generated impressions. After review, we summarize five types of errors in generated impressions, i.e., refuse-to-answer, truncated-output, repeated-output, no-output, and English-output errors. 

All four commercially available LLMs produce high-quality impressions with no truncated-output, repeated-output, or English-output errors. Only the Bard model refuses to provide answers for 4 PET-CT impression summarization prompts in one-shot and three-shot manners, respectively.

Different from the commercially available LLMs, the quality of generated impressions varies a lot among the four open-source LLMs. Figure \ref{fig3} shows the errors of the open-source LLMs. The baichuan model achieves high-quality results, where only one output has the no-output error. ChatGLM model also achieves good results when using zero-shot prompts, with only one truncated-output error. However, the ChatGLM model obtains many no-output errors when using few-shot prompts. Most of the no-output errors are due to the direct copy of the query section in the prompt but no generated impression. The two medical LLMs, HuatuoGPT and ChatGLM-Med, suffer serious errors in summarizing impressions. HuatuoGPT obtains truncated output and repeated-output errors in over 40\% of PET-CT impression summarization tasks. Although the percentage of errors in the CT and ultrasound impression summarization decreases, 13.67\% and 18.67\% of summarized impressions still contain repeated-output errors, respectively. Note that HuatuoGPT is more prone to obtain repeated output errors when using few-shot prompts. ChatGLM-Med obtains truncated output, repeated-output, and no-output errors in over 40\% of generated PET-CT impressions. And 13.33\% of generated CT impressions and 22.67\% of generated ultrasound impressions have truncated output, repeated output, and no output errors.

\begin{figure}[h]
\centering
\includegraphics[width=1.0\textwidth]{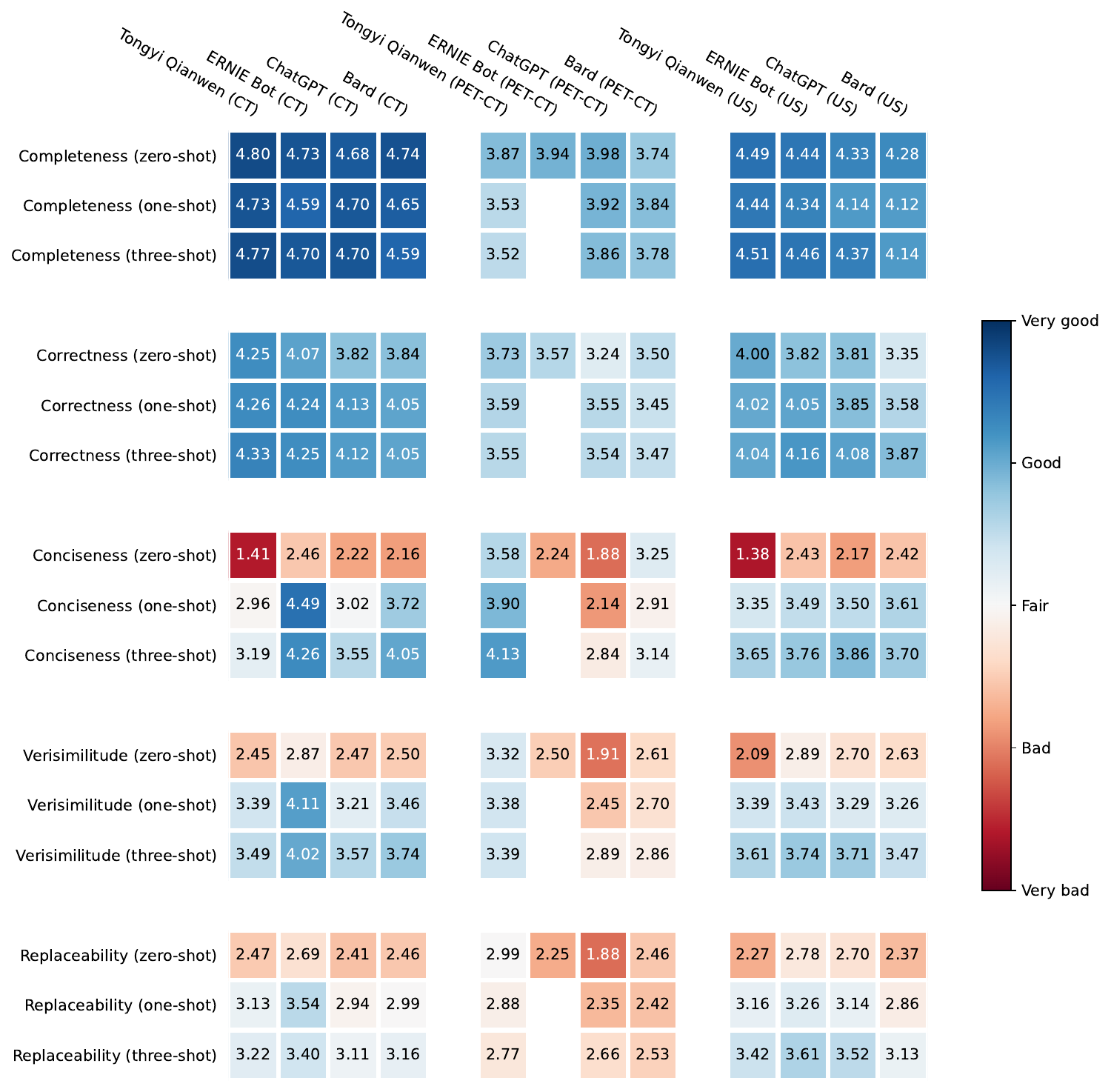}
\caption{Averaged human evaluation results of the generated impressions.}\label{fig4}
\end{figure}

\begin{table}[h]
\tiny
\caption{Averaged human evaluation results of the generated CT impressions.}\label{table4}
\begin{tabular*}{0.8\textwidth}{@{\extracolsep\fill}llcccc}

\toprule%
\multirow{2}{*}{Metric}              & \multirow{2}{*}{Prompt type}        & \multicolumn{4}{@{}c@{}}{Model}    \\
\cmidrule{3-6}
                                     &                    & Tongyi Qianwen & ERNIE Bot & ChatGPT & Bard   \\
\midrule
\multirow{3}{*}{Completeness}        & Zero-shot          & \textbf{4.80} & 4.73 & 4.68 & 4.74  \\
                                     & One-shot           & 4.73 & 4.59 & 4.70 & 4.65  \\
                                     & Three-shot         & 4.77 & 4.70 & 4.70 & 4.59\\
\cmidrule{2-6}
\multirow{3}{*}{Correctness}         & Zero-shot          & 4.25 & 4.07 & 3.82 & 3.84 \\
                                     & One-shot           & 4.26 & 4.24 & 4.13 & 4.05 \\
                                     & Three-shot         & \textbf{4.33} & 4.25 & 4.12 & 4.05\\
\cmidrule{2-6}
\multirow{3}{*}{Conciseness}         & Zero-shot          & 1.41 & 2.46 & 2.22 & 2.16\\
                                     & One-shot           & 2.96 & \textbf{4.49} & 3.02 & 3.72\\
                                     & Three-shot         & 3.19 & 4.26 & 3.55 & 4.05\\
\cmidrule{2-6}
\multirow{3}{*}{Verisimilitude}      & Zero-shot          & 2.45 & 2.87 & 2.47 & 2.50\\
                                     & One-shot           & 3.39 & \textbf{4.11} & 3.21 & 3.46\\
                                     & Three-shot         & 3.49 & 4.02 & 3.57 & 3.74\\
\cmidrule{2-6}
\multirow{3}{*}{Replaceability}      & Zero-shot          & 2.47 & 2.69 & 2.41 & 2.46\\
                                     & One-shot           & 3.13 & \textbf{3.54} & 2.94 & 2.99\\
                                     & Three-shot         & 3.22 & 3.40 & 3.11 & 3.16\\
\botrule
\end{tabular*}
\end{table}

\begin{table}[h]
\tiny
\caption{Averaged human evaluation results of the generated PET-CT impressions.}\label{table5}
\begin{tabular*}{0.8\textwidth}{@{\extracolsep\fill}llcccc}

\toprule%
\multirow{2}{*}{Metric}              & \multirow{2}{*}{Prompt type}        & \multicolumn{4}{@{}c@{}}{Model}    \\
\cmidrule{3-6}
                                     &                    & Tongyi Qianwen & ERNIE Bot & ChatGPT & Bard   \\
\midrule
\multirow{3}{*}{Completeness}        & Zero-shot          & 3.87 & 3.94 & \textbf{3.98} & 3.74\\
                                     & One-shot           & 3.53 & - & 3.92 & 3.84\\
                                     & Three-shot         & 3.52 & - & 3.86 & 3.78\\
\cmidrule{2-6}
\multirow{3}{*}{Correctness}         & Zero-shot          & \textbf{3.73} & 3.57 & 3.24 & 3.50\\
                                     & One-shot           & 3.59 & - & 3.55 & 3.45\\
                                     & Three-shot         & 3.55 & - & 3.54 & 3.47\\
\cmidrule{2-6}
\multirow{3}{*}{Conciseness}         & Zero-shot          & 3.58 & 2.24 & 1.88 & 3.25\\
                                     & One-shot           & 3.90 & - & 2.14 & 2.91\\
                                     & Three-shot         & \textbf{4.13} & - & 2.84 & 3.14\\
\cmidrule{2-6}
\multirow{3}{*}{Verisimilitude}      & Zero-shot          & 3.32 & 2.50 & 1.91 & 2.61\\
                                     & One-shot           & 3.38 & - & 2.45 & 2.70\\
                                     & Three-shot         & \textbf{3.39} & - & 2.89 & 2.86\\
\cmidrule{2-6}
\multirow{3}{*}{Replaceability}      & Zero-shot          & \textbf{2.99} & 2.25 & 1.88 & 2.46\\
                                     & One-shot           & 2.88 & - & 2.35 & 2.42\\
                                     & Three-shot         & 2.77 & - & 2.66 & 2.53\\
\botrule
\end{tabular*}
\end{table}

\begin{table}[h]
\tiny
\caption{Averaged human evaluation results of the generated Ultrasound impressions.}\label{table6}
\begin{tabular*}{0.8\textwidth}{@{\extracolsep\fill}llcccc}

\toprule%
\multirow{2}{*}{Metric}              & \multirow{2}{*}{Prompt type}        & \multicolumn{4}{@{}c@{}}{Model}    \\
\cmidrule{3-6}
                                     &                    & Tongyi Qianwen & ERNIE Bot & ChatGPT & Bard   \\
\midrule
\multirow{3}{*}{Completeness}        & Zero-shot          & 4.49 & 4.44 & 4.33 & 4.28\\
                                     & One-shot           & 4.44 & 4.34 & 4.14 & 4.12\\
                                     & Three-shot         & \textbf{4.51} & 4.46 & 4.37 & 4.14\\
\cmidrule{2-6}
\multirow{3}{*}{Correctness}         & Zero-shot          & 4.00 & 3.82 & 3.81 & 3.35\\
                                     & One-shot           & 4.02 & 4.05 & 3.85 & 3.58\\
                                     & Three-shot         & 4.04 & \textbf{4.16} & 4.08 & 3.87\\
\cmidrule{2-6}
\multirow{3}{*}{Conciseness}         & Zero-shot          & 1.38 & 2.43 & 2.17 & 2.42\\
                                     & One-shot           & 3.35 & 3.49 & 3.50 & 3.61\\
                                     & Three-shot         & 3.65 & 3.76 & \textbf{3.86} & 3.70\\
\cmidrule{2-6}
\multirow{3}{*}{Verisimilitude}      & Zero-shot          & 2.09 & 2.89 & 2.70 & 2.63\\
                                     & One-shot           & 3.39 & 3.43 & 3.29 & 3.26\\
                                     & Three-shot         & 3.61 & \textbf{3.74} & 3.71 & 3.47\\
\cmidrule{2-6}
\multirow{3}{*}{Replaceability}      & Zero-shot          & 2.27 & 2.78 & 2.70 & 2.37\\
                                     & One-shot           & 3.16 & 3.26 & 3.14 & 2.86\\
                                     & Three-shot         & 3.42 & \textbf{3.61} & 3.52 & 3.13\\
\botrule
\end{tabular*}
\end{table}

\subsubsection{Semantic evaluation}

Based on the automatic quantitative and manual quality evaluation, we note that the four commercially available LLMs achieve better impression summarization than the four open-source LLMs with higher BLEU, ROUGE-L, and METEOR values and better generation qualities. Therefore, we select the outputs of the four commercially available LLMs to further evaluate the semantics of the generated impressions. We define 5 human evaluation metrics: 1) Completeness, 2) Correctness, 3) Conciseness, 4) Verisimilitude, 5) Replaceability. The human evaluation results are shown in Tables \ref{table4}, \ref{table5}, and \ref{table6}. Figure \ref{fig4} illustrates the human evaluation results in a more intuitive way.

In the term of completeness, the generated CT and US impressions are significantly better than the generated PET-CT impressions. Clinical experts rate the generated CT and US impressions as between "Relatively complete" and "Very complete" (4.80 for CT and 4.51 for US), while the PET-CT impressions are only close to "Relatively complete" (3.98 for PET-CT). By comparing different prompt types, we note that the impressions obtained using zero-shot prompts achieve higher completeness score than the impressions obtained using few-shot prompts, but there is no significant differences in their completeness scores.

In the term of correctness, the generated CT and US impressions also achieve good results (4.33 for CT and 4.16 for US), which are between "Relatively correct" and "Very correct". The generated PET-CT impressions obtain the 3.73 for the correctness, not reaching the "Relatively correct" level. We also note that, when using few-shot prompts, the generated CT and US impressions get higher correctness scores, but lower correctness scores for the generated PET-CT impressions compared with using zero-shot prompts.

In the term of conciseness, the generated CT and PET-CT impressions obtain better results than the generated US impressions. Clinicians rate the generated CT and PET-CT impressions as between "Relatively concise" and "Very concise" (4.49 for CT and 4.13 for PET-CT), but the generated US impressions as between "Neutral" and "Relatively concise" (3.86 for US). When using few-shot prompts, the conciseness scores of the generated impressions achieve significant improvements compared with the generated impressions using zero-shot prompts.

In the term of verisimilitude, only the generated CT impressions score more than 4 point (4.11 for CT), while the generated PET-CT and US impressions score between "Neutral" and "Relatively verisimilar" (3.39 for PET-CT and 3.74 for US). Note that using few-shot prompts can also improve the verisimilitude of the generated impressions significantly.

To comprehensively evaluate the semantics of the generated impressions, clinical experts rate the replaceability of these impressions. We find that the impressions generated by LLMs are not yet at the level that can replace manually written impressions. The generated CT and US impressions only achieved replaceability scores of 3.54 and 3.61, which are between "Neutral" and "Relatively replaceable", while the generated PET-CT impressions have an even lower replaceability score of 2.99, which is only close to "Neutral".

When comparing the performances of different LLMs, we note that Tongyi Qianwen achieves the best results on the PET-CT impression generation task, with the best results in 4 of the 5 human evaluation metrics, i.e., correctness, conciseness, verisimilitude, and replaceability. ERNIE Bot outperforms the other LLMs on the CT impression generation task with the highest scores in conciseness, verisimilitude, and replaceability, and comparable scores in completeness and correctness. For US impression generation task, ERNIE Bot also achieves better results than other LLMs in correctness, verisimilitude, and replaceability and comparable results in completeness and conciseness. Note that the human evaluation results and the automatic quantitative evaluation results for the generated CT and PET-CT impressions are consistent, but not for the generated US impressions.

To analyze the evaluation variances between the clinical experts, we also list the evaluation results of each clinical expert in the Appendix Tables \ref{tableA1}, \ref{tableA2}, \ref{tableA3}, \ref{tableA4}, \ref{tableA5}, \ref{tableA6}, \ref{tableA7}, \ref{tableA8}, \ref{tableA9}, and Figure \ref{figA1}, \ref{figA2}, \ref{figA3}. Based on the results, we note that there are differences in the scores of different clinical experts. Clinician I's scores are relatively low. He thinks that none of the tree types of generated impressions can reach the level of replacing manually written impressions. Clinician II's scores are in the middle. He thinks that the generated CT and US impressions are close to replacing manually written impressions, but the generated PET-CT impressions are just neutral in the replaceability. Clinician III's scores are relatively high than the others. He thinks the generated PET-CT and CT impressions are close to replacing manually written impressions, and the generated US impressions can basically replace the manually written impressions.

Although the absolute values of the scores are different between clinical experts, the changing trends of impression scores under different prompt types are similar. Using few-shot prompts can improve most of the conciseness, verisimilitude, and replaceability scores significantly, but may lead to lower completeness and correctness scores. We also illustrate the significant test results in Appendix Figures \ref{figA4}, \ref{figA5}, \ref{figA6}, \ref{figA7}, \ref{figA8}, \ref{figA9}, \ref{figA10}, \ref{figA11}, \ref{figA12}.

\section{Discussion}

In this study, we aim to explore the current status of the LLMs in summarizing radiology report impressions. Automatic quantitative and human evaluations are conducted to measure the gap between the generated and reference impressions. 

\subsection{Commercially available LLMs vs open source LLMs}

To have a comprehensive evaluation of the state-of-the-art LLMs, in this study, we select four commercially available LLMs, i.e., ChatGPT, Bard, ERNIE Bot, and Tongyi Qianwen, and four open source LLMs, i.e., Baichuan, ChatGLM, HuatuoGPT, and ChatGLM-Med. According to the automatic quantitative evaluation, we can note that the commercially available LLMs outperform the open-source LLMs. Besides, the open-source LLMs exhibit more output errors in the generated impressions, such as the refuse-to-answer, truncated-output, repeated-output, no-output, and English-output errors. These errors are almost absent in the outputs of the commercially available LLMs. When using few-shot prompts, the commercially available LLMs can benefit more than the open-source LLMs, thus achieving higher improvements in the automatic quantitative evaluation metrics. The differences between the performance of commercially available and open-source LLMs may be due to the commercially available LLMs usually have more parameters, use more training data to train, employ more advanced closed-source algorithms to optimize, and are developed as web applications with better engineering implementations. The gap between the commercially available and open-source LLMs indicates that more computing resources and specialized engineering groups are critical for better LLMs, which has become the main obstacle for most research groups. 

\subsection{No best model for all impression summarization tasks}

In this study, we evaluate the LLMs under automatic quantitative and human evaluation metrics. Based on the evaluation results, no single LLM can achieve the best results in all impression summarization tasks. Using automatic quantitative evaluation metrics, Tongyi Qianwen, ERNIE Bot, and ChatGPT achieve the best overall performance in the PET-CT, CT, and US impression summarization tasks, respectively. When evaluated by clinical experts, Tongyi Qianwen and ERNIE Bot are the best LLMs for the PET-CT and CT impression summarization tasks, respectively. But for US impression summarization, the clinical experts think the ERNIE Bot is better than ChatGPT. Although the experimental results indicate the evaluated LLMs are very competitive with each other and no one can outperform others in all impression summarization tasks significantly, we note that the Chinese LLMs achieve almost all the best results for automatic quantitative and human evaluation metrics except the ChatGPT for US impression summarization under automatic quantitative evaluation. This finding suggests the necessity to build LLMs for specific languages, which can achieve better performance on language-specific tasks.

\subsection{Effect of the few-shot prompt}

In this study, we also explore the effect of the few-shot prompt on impression summarization. Based on the experimental results, we note that the few-shot prompt can significantly improve the performance of LLMs on all automatic quantitative evaluation metrics and some human evaluation metrics, including conciseness, verisimilitude, and replaceability. For correctness and completeness, using few-shot prompts may lead to some performance degradation, but usually not significant. When further comparing the performance of LLMs using one-shot and three-shot prompts, we find that the more examples provided, the better the impressions generated are not achieved. For example, Tongyi Qianwen achieves the best BLEU values when using one-shot prompts and the best ROUGE-L and METEOR values when using three-shot prompts for PET-CT impression summarization. ERINE Bot outperforms the other LLMs in ROUGE-L for CT impression summarization when using one-shot prompts but achieves the best ROUGE-L scores for US impression summarization when using three-shot prompts. Although there is an overall trend that using few-shot prompts will improve the performance of LLMs in generating impressions, it seems unclear how many examples a prompt should include to be most effective. 

\subsection{Clinical application}

Note that to evaluate the semantics of the generated impressions, we first extract the impressions from the generated text manually and then conduct the human evaluation. Therefore, the current experimental results may be higher than those obtained by evaluating the original outputs. We list the automatic quantitative results in Appendix Tables \ref{tableA10}, \ref{tableA11}, \ref{tableA12} and Figure \ref{figA13}. We also show the difference in results between using the extracted impressions and original outputs in Figures \ref{figA14} and \ref{figA15}. We can note that all results obtained by evaluating extracted impressions are higher than those on original outputs. However, among all LLMs, Tongyi Qianwen, ERNIE Bot, and ChatGPT show small differences between these results, indicating they can follow the instructions well to generate the text we desire. Although the Bard achieves comparable performance based on the extracted impressions, its original outputs contain much more impression-unrelated content, reducing its usability in summarizing impressions in real clinical practice.

According to the evaluation of clinical experts, the impressions generated by the LLMs can not directly replace the impressions written by radiologists. However, using LLMs to summarize clinical text like radiology findings is still valuable. First, it can help clinicians improve the efficiency of writing clinical documents. In clinical practice, writing clinical documents like radiology impressions, admission records, progress notes, and discharge summaries is time-consuming and tedious. To alleviate this problem, we can use the LLMs to summarize the related structured or unstructured electronic health records as a preliminary clinical note, and then the clinicians conduct the final review. For cancer patients who usually undergo a long diagnosis and treatment process, we can also employ the LLMs to summarize the whole diagnosis and treatment timeline, which is very important and valuable for the development of the next treatment plan. Second, LLMs may improve the diagnostic capabilities of primary care physicians. There are significant differences in the ability of physicians to diagnose benign findings associated with tumors. Primary care physicians who lack rich experience are more likely to overdiagnose. LLMs can generate the impressions to assist primary care physicians in making more accurate diagnoses. Thirdly, we can use LLMs to facilitate the research. Based on the summarization ability, LLMs can effectively extract key information from clinical documents to identify eligible patients for specific studies.

\subsection{Limitations and future work}

To comprehensively evaluate the LLMs' impression summarizing ability, we select three types of radiology reports, i.e., PET-CT, CT, and Ultrasound reports. We should note that all reports are from lung cancer patients treated in a single medical center, which indicates the patient population is homogeneous and the writing style of the reports is relatively uniform. So, the results in this study may differ from the average performance of LLMs in summarizing the impressions of reports from patients with different diseases or medical centers. In the future, we will try to collect more radiology reports from different patients and medical centers to evaluate the LLMs to obtain more robust results. 

The most important contribution of this study is that we invite three clinical experts to manually evaluate the impressions generated by the LLMs from the point of view of semantics so that we can find out the gap between the reports generated by LLMs and those written by radiologists. However, manual evaluation is time-consuming and tedious, which is the biggest obstacle for a large amount of evaluation. Therefore, in this study, we only recruit three clinical experts, i.e., two thoracic surgeons and one radiologist, to evaluate 100 generated impressions for each type of report. To obtain more convincing results, we will try to recruit more clinicians with different years of experience from different departments to evaluate more impressions in the future.

Currently, LLMs are updated very quickly. Since human evaluation is very time-consuming, we can not perform real-time human evaluation of the latest LLMs. In the future, we will try to evaluate the latest LLMs and compare them with their previous versions to find out the changes in the performance of impression summarization of radiology reports.

\section{Methods}

\subsection{Materials}

We collected three types of radiology reports, i.e., PET-CT, CT, and ultrasound (US) reports from Peking University Cancer Hospital and Institute. The relevant patients are all outpatients and inpatients of the Department of Thoracic Surgery II. After removing the incomplete reports, we finally obtain 867 PET-CT reports, 819 CT reports, and 1487 ultrasound reports. We randomly select 100 reports from each type of report for automatic quantitative and human evaluations.

\subsection{Large language models}

In this study, we aim to explore the current status of prompt-based LLMs in radiology report impression summarization. To conduct a comprehensive evaluation, we select four commercially available and four open source LLMs with different architectures and parameter sizes. The summaries of the selected LLMs are listed below.
\begin{itemize}

\item \textbf{Tongyi Qianwen}. Tongyi Qianwen is an LLM chat product developed by Alibaba Cloud. The latest Tongyi Qianwen 2.0 extends the Qwen model \cite{Bai2023qwen} to a few hundred billion parameters, achieving a substantial upgrade from its predecessor in understanding complex instructions, reasoning, memorizing, and preventing hallucinations. We used Tongyi Qianwen v2.1.1 (\url{https://tongyi.aliyun.com/qianwen/}) to generate the impressions.

\item \textbf{ERNIE Bot}. ERNIE Bot (Wenxin Yiyan) is an LLM chat product developed by Baidu based on their ERNIE (Enhanced Representation through Knowledge Integration) \cite{Sun2019} and PLATO (Pre-trained Dialogue Generation Model) \cite{Bao2020} models. Based on the supervised fine-tuning, RLHF, and knowledge, search, dialogue enhancements, the ERNIE Bot achieves a more precise understanding of Chinese language and its practical applications. We used ERNIE Bot v2.5.2 (\url{https://yiyan.baidu.com/}) to generate the impressions.

\item \textbf{ChatGPT}. ChatGPT is the most impactful LLM developed by OpenAI, raising the trend of prompt-based LLMs worldwide. ChatGPT is an advanced version of instructionGPT \cite{Ouyang2022}, which first fine-tunes GPT-3 \cite{Brown2020} using human-written demonstrations of the desired output to prompts and then further fine-tuning the model through the RLHF strategy to align language models with user intent. We accessed the ChatGPT via website interface (\url{https://chatgpt.com/}) to obtain the generated impressions before January 11, 2024.

\item \textbf{Bard}. Bard is an LLM chat product powered by PaLM 2 \cite{Anil2023} (Pathways Language Model 2) developed by Google AI. PaLM 2 is a transformer-based model trained using a mixture of objectives and multilingual datasets, achieving better performances on natural language generation, code generation, translation, and reasoning than its predecessor, PaLM \cite{Aakanksha2023}. We accessed the Bard via website interface (\url{https://gemini.google.com/app}) to obtain the generated impressions before January 12, 2024.

\item \textbf{Baichuan}. Baichuan-13B \cite{baichuan2023} is an open-source LLM developed by Baichuan Intelligence. The baichuan-13B model has 130 billion parameters trained on 1.4 trillion tokens. It supports both Chinese and English and achieves competitive performance in standard Chinese and English benchmarks among models of its size. We used the Baichuan-13B-Chat (\url{https://huggingface.co/baichuan-inc/Baichuan-13B-Chat}) to generate the impressions.

\item \textbf{ChatGLM}. ChatGLM3-6B is the latest open-source model in the ChatGLM \cite{Du2022} series developed by Tsinghua University. The ChatGLM3-6B has a more powerful base model trained on a more diverse dataset,  sufficient training steps, and a more reasonable training strategy, showing strong performance on language understanding, reasoning, coding, etc. We used the ChatGLM3-6b (\url{https://huggingface.co/THUDM/chatglm3-6b}) to generated the impressions.

\item \textbf{HuatuoGPT}. HuatuoGPT \cite{Zhang2023huatuogpt} is an open-source LLM developed by the Shenzhen Research Institute of Big Data. HuatuoGPT-7B first uses the Baichuan-7B as the backbone model and then uses the distilled data from ChatGPT and real-world data from doctors to supervised fine-tuning and reinforcement learning with mixed feedback to achieve state-of-the-art results in performing medical consultation. We used the HuatuoGPT-7B (\url{https://github.com/FreedomIntelligence/HuatuoGPT}) to generate the impressions.

\item \textbf{ChatGLM-Med}. ChatGLM-Med \cite{Wang2023} is an open-source LLM developed by the Harbin Institution of Technology. The ChatGLM-Med employs the ChatGLM-6B as the base model and fine-tunes on a Chinese medical instruction dataset developed by a medical knowledge graph and GPT-3.5 to improve better question-answering results in the medical field. We used the ChatGLM-Med (\url{https://github.com/SCIR-HI/Med-ChatGLM}) to generate the impressions.

\end{itemize}

\begin{figure}[h]
\centering
\includegraphics[width=1.0\textwidth]{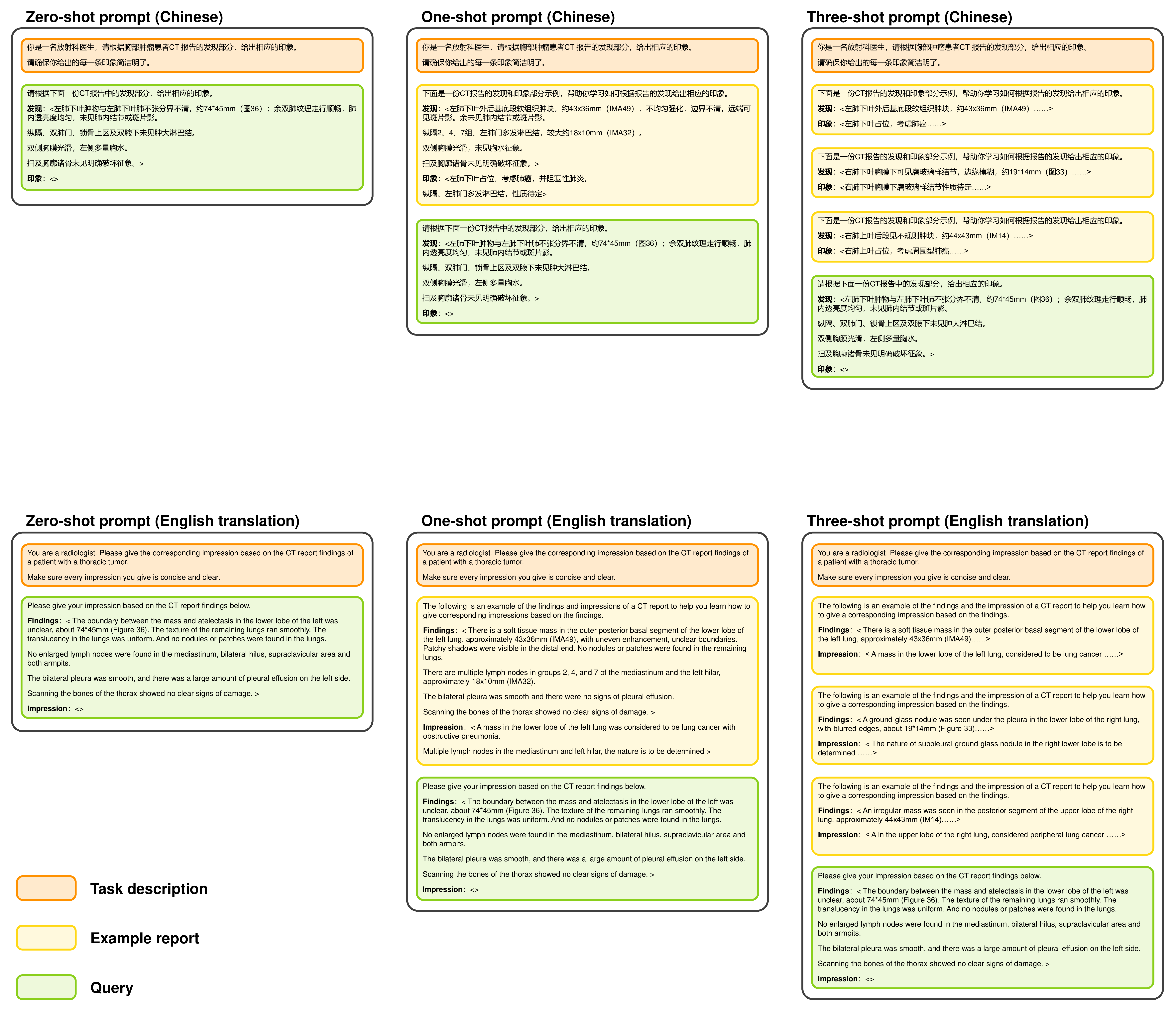}
\caption{Zero-shot, one-shot and three-shot prompts in Chinese and English.}\label{fig5}
\end{figure}

\subsection{Impression summarization using LLMs}

To explore the capability of LLMs to summarize the impression in a zero-shot or few-shot manner, we first design the zero, one, and three-shot prompts as shown in Figure \ref{fig5}. The zero-shot prompt consists of two parts, i.e., task description and query. The one-shot and three-shot prompts add one and three example reports between the task description and query, respectively. Note that the example reports are randomly selected from the dataset and different from the report in the query. 

Using the developed prompts, we collect the outputs of the four commercially available LLMs from their corresponding websites manually. And we deploy the four open-source LLMs on our server to obtain their outputs. Note that, besides the summarized impression, the LLMs usually generate some other content such as the findings, the future examination advice, the explanation of the response, etc. To accurately evaluate the generated impressions, we conduct a post-processing procedure to remove the unrelated content from the outputs to keep the impressions only for further quantitative and human evaluations.

\subsection{Quantitative evaluation metrics}

In this study, we select 3 metrics widely used in text generation research to evaluate the generated impression against the reference impression. The values of these metrics range from 0 to 1, where a higher value indicates a better result. A brief introduction of the metrics is listed below:
% We use the Student T test to compare the quantitative evaluation metrics of different LLMs with different prompts.

\begin{itemize}
\item \textbf{BLEU}. BiLingual Evaluation Understudy (BLEU) \cite{Papineni2002} score measures the number of position-independent matches of the n-grams of the candidate with the n-grams of the reference, focusing on the precision of the n-grams.

\item \textbf{ROUGE-L}. Recall-Oriented Understudy for Gisting Evaluation using Longest Common Subsequence (ROUGE-L) \cite{Lin2004} measures the longest common subsequence (LCS) between the candidate and reference to calculate the LCS-based F-measure.

\item \textbf{METEOR}. Metric for Evaluation of Translation with Explicit ORdering (METEOR) \cite{Banerjee2005} measures the harmonic mean of precision and recall calculated based on the mapping between unigrams with the least number of crosses by exact, stemming, and synonym matching.
\end{itemize}

\subsection{Human evaluation metrics}

Although the automatic quantitative evaluation metrics above have shown some correlations with human judgments, they are not sufficient enough to evaluate the difference between the generated and reference impressions in semantics. Therefore, we define 5 human evaluation metrics, i.e., 1) Correctness, 2) Completeness, 3) Conciseness, 4) Verisimilitude, and 5) Replaceability, in this study to evaluate the semantics of the generated impressions. The definitions are listed below. We recruit three clinical experts (ZSY, LB, and LQ) to annotate the generated impression. We use a 5-point Likert scale for each evaluation metric. A higher value indicates a better result. Note that the clinical experts are blinded to the LLM and prompt types when annotating. We use the Mann-Whitney U test to compare the human evaluation metrics of different LLMs with different prompts.

\begin{itemize}

\item \textbf{Completeness}. Completeness measures how completely the information in the generated impression covers the information in the reference impression. The five answer statements for the 5-point Likert scale of Completeness are 1 for "Very incomplete", 2 for "Relatively incomplete", 3 for "Neutral", 4 for "Relatively completeness", and 5 for "Very correct".

\item \textbf{Correctness}. Correctness measures how correct the information in the generated impression is compared to the information in the reference impression. The five answer statements for the 5-point Likert scale of Correctness are 1 for "Very incorrect", 2 for "Relatively incorrect", 3 for "Neutral", 4 for "Relatively correct", and 5 for "Very correct".

\item \textbf{Conciseness}. Conciseness measures how much redundant information is in the generated impression. The five answer statements for the 5-point Likert scale of Conciseness are 1 for "Very redundant", 2 for "Relatively redundant", 3 for "Neutral", 4 for "Relatively concise", and 5 for "Very concise".

\item \textbf{Verisimilitude}. Verisimilitude measures how similar the generated impression is to the reference impression in readability, grammar, and writing style. The five answer statements for the 5-point Likert scale of Verisimilitude are 1 for "Very fake", 2 for "Relatively fake", 3 for "Neutral", 4 for "Relatively verisimilar", and 5 for "Very verisimilar".

\item \textbf{Replaceability}. Replaceability measures whether the generated impression can replace the reference impression. The five answer statements for the 5-point Likert scale of Replaceability are 1 for "Very irreplaceable", 2 for "Relatively irreplaceable", 3 for "Neutral", 4 for "Relatively replaceable", and 5 for "Very replaceable".

\end{itemize}

\section{Conclusion}

In this study, we explore the current status of LLMs in summarizing radiology report impressions using automatic quantitative and human evaluations. The experimental results indicate that there is a gap between the impressions generated by LLMs and written by radiologists. LLMs achieve great performance in completeness and correctness, but are not good in conciseness and verisimilitude. Although the few-shot prompt can improve the LLMs' performance in conciseness and verisimilitude, clinicians still believe that the LLMs can not replace the radiologist in summarizing impressions, especially for PET-CT reports with long findings.

\backmatter

% \bmhead{Supplementary information}

% If your article has accompanying supplementary file/s please state so here. 

% Authors reporting data from electrophoretic gels and blots should supply the full unprocessed scans for key as part of their Supplementary information. This may be requested by the editorial team/s if it is missing.

% Please refer to Journal-level guidance for any specific requirements.

\bmhead{Acknowledgements}

This work was supported by the Beijing Natural Science Foundation (L222020), the National Key R\&D Program of China (No.2022YFC2406804), the Capital’s funds for health improvement and research (No.2024-1-1023), and the National Ten-thousand Talent Program.

% \section*{Declarations}

% Some journals require declarations to be submitted in a standardised format. Please check the Instructions for Authors of the journal to which you are submitting to see if you need to complete this section. If yes, your manuscript must contain the following sections under the heading `Declarations':

% \begin{itemize}
% \item Conflict of interest/Competing interests (check journal-specific guidelines for which heading to use)
% \item Ethics approval and consent to participate
% \item Consent for publication
% \item Materials availability
% \end{itemize}

% \noindent
% If any of the sections are not relevant to your manuscript, please include the heading and write `Not applicable' for that section. 

\bibliography{sn-bibliography}% common bib file

\begin{appendices}

\section{}\label{secA1}




% \section{Example of another appendix section}\label{secA2}%

\begin{table}[h]
\tiny
\caption{Clinician I's evaluation results of the generated CT impressions.}\label{tableA1}
\begin{tabular*}{0.8\textwidth}{@{\extracolsep\fill}llcccc}

\toprule%
\multirow{2}{*}{Metric}              & \multirow{2}{*}{Prompt type}        & \multicolumn{4}{@{}c@{}}{Model}    \\
\cmidrule{3-6}
                                     &                    & Tongyi Qianwen & ERNIE Bot & ChatGPT & Bard   \\
\midrule
\multirow{3}{*}{Completeness}        & Zero-shot          & \textbf{4.90} & 4.83 & 4.80 & 4.87\\
                                     & One-shot           & 4.84 & 4.66 & 4.79 & 4.77\\
                                     & Three-shot         & 4.85 & 4.75 & 4.77 & 4.70\\
\cmidrule{2-6}
\multirow{3}{*}{Correctness}         & Zero-shot          & 4.57 & 4.61 & 4.41 & 4.53\\
                                     & One-shot           & 4.52 & 4.54 & 4.44 & 4.55\\
                                     & Three-shot         & \textbf{4.64} & 4.54 & 4.45 & 4.44\\
\cmidrule{2-6}
\multirow{3}{*}{Conciseness}         & Zero-shot          & 1.32 & 2.24 & 1.90 & 1.90\\
                                     & One-shot           & 2.87 & \textbf{4.38} & 2.56 & 3.59\\
                                     & Three-shot         & 2.95 & 3.99 & 3.10 & 3.81\\
\cmidrule{2-6}
\multirow{3}{*}{Verisimilitude}      & Zero-shot          & 1.96 & 2.47 & 2.01 & 2.22\\
                                     & One-shot           & 3.21 & \textbf{4.03} & 2.66 & 3.36\\
                                     & Three-shot         & 3.15 & 3.81 & 3.13 & 3.52\\
\cmidrule{2-6}
\multirow{3}{*}{Replaceability}      & Zero-shot          & 2.23 & 2.14 & 1.84 & 2.04\\
                                     & One-shot           & 2.64 & \textbf{2.91} & 2.16 & 2.31\\
                                     & Three-shot         & 2.63 & 2.58 & 2.26 & 2.39\\
\botrule
\end{tabular*}
\end{table}

\begin{table}[h]
\tiny
\caption{Clinician II's evaluation results of the generated CT impressions.}\label{tableA2}
\begin{tabular*}{0.8\textwidth}{@{\extracolsep\fill}llcccc}

\toprule%
\multirow{2}{*}{Metric}              & \multirow{2}{*}{Prompt type}        & \multicolumn{4}{@{}c@{}}{Model}    \\
\cmidrule{3-6}
                                     &                    & Tongyi Qianwen & ERNIE Bot & ChatGPT & Bard   \\
\midrule
\multirow{3}{*}{Completeness}        & Zero-shot          & \textbf{4.51} & 4.36 & 4.38 & 4.35\\
                                     & One-shot           & 4.40 & 4.30 & 4.41 & 4.24\\
                                     & Three-shot         & 4.50 & 4.45 & 4.42 & 4.22\\
\cmidrule{2-6}
\multirow{3}{*}{Correctness}         & Zero-shot          & 3.77 & 3.62 & 4.03 & 3.27\\
                                     & One-shot           & 4.01 & 4.12 & 4.12 & 3.73\\
                                     & Three-shot         & 4.08 & 4.12 & \textbf{4.17} & 3.73\\
\cmidrule{2-6}
\multirow{3}{*}{Conciseness}         & Zero-shot          & 1.35 & 2.49 & 2.57 & 2.29\\
                                     & One-shot           & 2.93 & \textbf{4.44} & 3.36 & 3.74\\
                                     & Three-shot         & 3.30 & 4.31 & 3.77 & 4.02\\
\cmidrule{2-6}
\multirow{3}{*}{Verisimilitude}      & Zero-shot          & 2.22 & 2.74 & 2.88 & 2.49\\
                                     & One-shot           & 3.18 & 3.85 & 3.42 & 3.11\\
                                     & Three-shot         & 3.43 & \textbf{3.89} & 3.65 & 3.45\\
\cmidrule{2-6}
\multirow{3}{*}{Replaceability}      & Zero-shot          & 1.83 & 2.72 & 2.94 & 2.45\\
                                     & One-shot           & 3.14 & \textbf{3.99} & 3.44 & 3.27\\
                                     & Three-shot         & 3.46 & 3.87 & 3.82 & 3.56\\
\botrule
\end{tabular*}
\end{table}

\begin{table}[h]
\tiny
\caption{Clinician III's evaluation results of the generated CT impressions.}\label{tableA3}
\begin{tabular*}{0.8\textwidth}{@{\extracolsep\fill}llcccc}

\toprule%
\multirow{2}{*}{Metric}              & \multirow{2}{*}{Prompt type}        & \multicolumn{4}{@{}c@{}}{Model}    \\
\cmidrule{3-6}
                                     &                    & Tongyi Qianwen & ERNIE Bot & ChatGPT & Bard   \\
\midrule
\multirow{3}{*}{Completeness}        & Zero-shot          & 4.98 & \textbf{4.99} & 4.87 & \textbf{4.99}\\
                                     & One-shot           & 4.95 & 4.81 & 4.90 & 4.95\\
                                     & Three-shot         & 4.95 & 4.89 & 4.92 & 4.85\\
\cmidrule{2-6}
\multirow{3}{*}{Correctness}         & Zero-shot          & \textbf{4.40} & 3.97 & 3.01 & 3.73\\
                                     & One-shot           & 4.24 & 4.07 & 3.83 & 3.87\\
                                     & Three-shot         & 4.27 & 4.08 & 3.75 & 3.99\\
\cmidrule{2-6}
\multirow{3}{*}{Conciseness}         & Zero-shot          & 1.57 & 2.66 & 2.18 & 2.29\\
                                     & One-shot           & 3.09 & \textbf{4.66} & 3.13 & 3.82\\
                                     & Three-shot         & 3.32 & 4.47 & 3.79 & 4.31\\
\cmidrule{2-6}
\multirow{3}{*}{Verisimilitude}      & Zero-shot          & 3.17 & 3.40 & 2.51 & 2.78\\
                                     & One-shot           & 3.77 & \textbf{4.44} & 3.54 & 3.91\\
                                     & Three-shot         & 3.89 & 4.35 & 3.93 & 4.24\\
\cmidrule{2-6}
\multirow{3}{*}{Replaceability}      & Zero-shot          & 3.35 & 3.22 & 2.44 & 2.89\\
                                     & One-shot           & 3.61 & 3.72 & 3.21 & 3.40\\
                                     & Three-shot         & 3.58 & \textbf{3.74} & 3.24 & 3.54\\
\botrule
\end{tabular*}
\end{table}

\begin{table}[h]
\tiny
\caption{Clinician I's evaluation results of the generated PET-CT impressions.}\label{tableA4}
\begin{tabular*}{0.8\textwidth}{@{\extracolsep\fill}llcccc}

\toprule%
\multirow{2}{*}{Metric}              & \multirow{2}{*}{Prompt type}        & \multicolumn{4}{@{}c@{}}{Model}    \\
\cmidrule{3-6}
                                     &                    & Tongyi Qianwen & ERNIE Bot & ChatGPT & Bard   \\
\midrule
\multirow{3}{*}{Completeness}        & Zero-shot          & 3.81 & 3.94 & \textbf{4.07} & 3.86\\
                                     & One-shot           & 3.43 & - & 3.81 & 3.88\\
                                     & Three-shot         & 3.49 & - & 3.74 & 3.74\\
\cmidrule{2-6}
\multirow{3}{*}{Correctness}         & Zero-shot          & 3.70 & 3.64 & \textbf{3.79} & 3.68\\
                                     & One-shot           & 3.61 & - & 3.64 & 3.68\\
                                     & Three-shot         & 3.66 & - & 3.56 & 3.57\\
\cmidrule{2-6}
\multirow{3}{*}{Conciseness}         & Zero-shot          & 2.87 & 1.76 & 1.37 & 2.73\\
                                     & One-shot           & 3.59 & - & 1.47 & 2.20\\
                                     & Three-shot         & \textbf{3.75} & - & 1.98 & 2.56\\
\cmidrule{2-6}
\multirow{3}{*}{Verisimilitude}      & Zero-shot          & 2.97 & 2.22 & 1.42 & 2.22\\
                                     & One-shot           & 3.41 & - & 1.94 & 2.33\\
                                     & Three-shot         & \textbf{3.46} & - & 2.33 & 2.63\\
\cmidrule{2-6}
\multirow{3}{*}{Replaceability}      & Zero-shot          & 1.97 & 1.63 & 1.24 & 1.67\\
                                     & One-shot           & \textbf{1.99} & - & 1.60 & 1.58\\
                                     & Three-shot         & 1.93 & - & 1.70 & 1.66\\
\botrule
\end{tabular*}
\end{table}

\begin{table}[h]
\tiny
\caption{Clinician II's evaluation results of the generated PET-CT impressions.}\label{tableA5}
\begin{tabular*}{0.8\textwidth}{@{\extracolsep\fill}llcccc}

\toprule%
\multirow{2}{*}{Metric}              & \multirow{2}{*}{Prompt type}        & \multicolumn{4}{@{}c@{}}{Model}    \\
\cmidrule{3-6}
                                     &                    & Tongyi Qianwen & ERNIE Bot & ChatGPT & Bard   \\
\midrule
\multirow{3}{*}{Completeness}        & Zero-shot          & 3.73 & 3.87 & 3.89 & 3.49\\
                                     & One-shot           & 3.40 & - & \textbf{3.94} & 3.66\\
                                     & Three-shot         & 3.18 & - & 3.81 & 3.61\\
\cmidrule{2-6}
\multirow{3}{*}{Correctness}         & Zero-shot          & \textbf{3.49} & 3.20 & 3.34 & 2.88\\
                                     & One-shot           & 3.23 & - & 3.17 & 2.68\\
                                     & Three-shot         & 3.03 & - & 3.21 & 2.89\\
\cmidrule{2-6}
\multirow{3}{*}{Conciseness}         & Zero-shot          & 3.61 & 2.22 & 1.92 & 3.57\\
                                     & One-shot           & 3.97 & - & 2.02 & 3.02\\
                                     & Three-shot         & \textbf{4.09} & - & 2.69 & 3.20\\
\cmidrule{2-6}
\multirow{3}{*}{Verisimilitude}      & Zero-shot          & \textbf{2.96} & 2.32 & 2.06 & 2.44\\
                                     & One-shot           & 2.77 & - & 2.16 & 2.31\\
                                     & Three-shot         & 2.66 & - & 2.48 & 2.46\\
\cmidrule{2-6}
\multirow{3}{*}{Replaceability}      & Zero-shot          & \textbf{3.02} & 2.32 & 2.08 & 2.54\\
                                     & One-shot           & 2.77 & - & 2.21 & 2.25\\
                                     & Three-shot         & 2.56 & - & 2.54 & 2.48\\
\botrule
\end{tabular*}
\end{table}

\begin{table}[h]
\tiny
\caption{Clinician III's evaluation results of the generated PET-CT impressions.}\label{tableA6}
\begin{tabular*}{0.8\textwidth}{@{\extracolsep\fill}llcccc}

\toprule%
\multirow{2}{*}{Metric}              & \multirow{2}{*}{Prompt type}        & \multicolumn{4}{@{}c@{}}{Model}    \\
\cmidrule{3-6}
                                     &                    & Tongyi Qianwen & ERNIE Bot & ChatGPT & Bard   \\
\midrule
\multirow{3}{*}{Completeness}        & Zero-shot          & \textbf{4.07} & 4.01 & 3.99 & 3.88\\
                                     & One-shot           & 3.77 & - & 4.02 & 4.00\\
                                     & Three-shot         & 3.88 & - & 4.04 & 4.00\\
\cmidrule{2-6}
\multirow{3}{*}{Correctness}         & Zero-shot          & \textbf{4.01} & 3.86 & 2.58 & 3.95\\
                                     & One-shot           & 3.92 & - & 3.84 & 4.00\\
                                     & Three-shot         & 3.96 & - & 3.84 & 3.96\\
\cmidrule{2-6}
\multirow{3}{*}{Conciseness}         & Zero-shot          & 4.25 & 2.73 & 2.34 & 3.46\\
                                     & One-shot           & 4.13 & - & 2.93 & 3.51\\
                                     & Three-shot         & \textbf{4.56} & - & 3.84 & 3.67\\
\cmidrule{2-6}
\multirow{3}{*}{Verisimilitude}      & Zero-shot          & 4.03 & 2.96 & 2.25 & 3.17\\
                                     & One-shot           & 3.96 & - & 3.26 & 3.45\\
                                     & Three-shot         & \textbf{4.06} & - & 3.85 & 3.51\\
\cmidrule{2-6}
\multirow{3}{*}{Replaceability}      & Zero-shot          & \textbf{3.98} & 2.81 & 2.32 & 3.17\\
                                     & One-shot           & 3.89 & - & 3.25 & 3.42\\
                                     & Three-shot         & 3.82 & - & 3.73 & 3.47\\
\botrule
\end{tabular*}
\end{table}

\begin{table}[h]
\tiny
\caption{Clinician I's evaluation results of the generated Ultrasound impressions.}\label{tableA7}
\begin{tabular*}{0.8\textwidth}{@{\extracolsep\fill}llcccc}

\toprule%
\multirow{2}{*}{Metric}              & \multirow{2}{*}{Prompt type}        & \multicolumn{4}{@{}c@{}}{Model}    \\
\cmidrule{3-6}
                                     &                    & Tongyi Qianwen & ERNIE Bot & ChatGPT & Bard   \\
\midrule
\multirow{3}{*}{Completeness}        & Zero-shot          & 4.11 & 4.01 & 3.82 & 3.88\\
                                     & One-shot           & 4.10 & 4.01 & 3.67 & 3.75\\
                                     & Three-shot         & \textbf{4.20} & 4.08 & 3.95 & 3.71\\
\cmidrule{2-6}
\multirow{3}{*}{Correctness}         & Zero-shot          & 3.57 & 3.53 & 3.56 & 3.35\\
                                     & One-shot           & 3.66 & 3.77 & 3.47 & 3.49\\
                                     & Three-shot         & 3.71 & \textbf{3.88} & 3.63 & 3.57\\
\cmidrule{2-6}
\multirow{3}{*}{Conciseness}         & Zero-shot          & 1.05 & 1.77 & 1.38 & 1.69\\
                                     & One-shot           & 2.76 & 2.85 & 2.80 & 2.91\\
                                     & Three-shot         & 3.00 & \textbf{3.09} & \textbf{3.09} & 2.98\\
\cmidrule{2-6}
\multirow{3}{*}{Verisimilitude}      & Zero-shot          & 1.41 & 2.09 & 1.69 & 1.97\\
                                     & One-shot           & 2.72 & 2.80 & 2.34 & 2.76\\
                                     & Three-shot         & 2.86 & \textbf{3.05} & 2.98 & 2.83\\
\cmidrule{2-6}
\multirow{3}{*}{Replaceability}      & Zero-shot          & 1.66 & 1.80 & 1.77 & 1.32\\
                                     & One-shot           & 2.23 & 2.21 & 2.10 & 1.84\\
                                     & Three-shot         & 2.34 & \textbf{2.57} & 2.42 & 1.98\\
\botrule
\end{tabular*}
\end{table}

\begin{table}[h]
\tiny
\caption{Clinician II's evaluation results of the generated Ultrasound impressions.}\label{tableA8}
\begin{tabular*}{0.8\textwidth}{@{\extracolsep\fill}llcccc}

\toprule%
\multirow{2}{*}{Metric}              & \multirow{2}{*}{Prompt type}        & \multicolumn{4}{@{}c@{}}{Model}    \\
\cmidrule{3-6}
                                     &                    & Tongyi Qianwen & ERNIE Bot & ChatGPT & Bard   \\
\midrule
\multirow{3}{*}{Completeness}        & Zero-shot          & 4.36 & 4.32 & 4.17 & 3.96\\
                                     & One-shot           & 4.22 & 4.01 & 3.75 & 3.64\\
                                     & Three-shot         & \textbf{4.37} & 4.29 & 4.16 & 3.73\\
\cmidrule{2-6}
\multirow{3}{*}{Correctness}         & Zero-shot          & 4.03 & 3.85 & 3.75 & 2.54\\
                                     & One-shot           & 3.96 & 3.88 & 3.77 & 2.96\\
                                     & Three-shot         & 4.04 & \textbf{4.14} & 3.95 & 3.31\\
\cmidrule{2-6}
\multirow{3}{*}{Conciseness}         & Zero-shot          & 1.09 & 2.40 & 2.02 & 2.36\\
                                     & One-shot           & 3.21 & 3.44 & 3.56 & 3.67\\
                                     & Three-shot         & 3.58 & 3.81 & \textbf{3.88} & 3.77\\
\cmidrule{2-6}
\multirow{3}{*}{Verisimilitude}      & Zero-shot          & 1.85 & 2.57 & 2.34 & 1.86\\
                                     & One-shot           & 3.14 & 3.13 & 3.13 & 2.62\\
                                     & Three-shot         & 3.54 & \textbf{3.71} & 3.51 & 2.92\\
\cmidrule{2-6}
\multirow{3}{*}{Replaceability}      & Zero-shot          & 1.33 & 2.54 & 2.29 & 1.75\\
                                     & One-shot           & 3.13 & 3.25 & 3.11 & 2.57\\
                                     & Three-shot         & 3.67 & \textbf{3.82} & 3.59 & 2.93\\
\botrule
\end{tabular*}
\end{table}

\begin{table}[h]
\tiny
\caption{Clinician III's evaluation results of the generated Ultrasound impressions.}\label{tableA9}
\begin{tabular*}{0.8\textwidth}{@{\extracolsep\fill}llcccc}

\toprule%
\multirow{2}{*}{Metric}              & \multirow{2}{*}{Prompt type}        & \multicolumn{4}{@{}c@{}}{Model}    \\
\cmidrule{3-6}
                                     &                    & Tongyi Qianwen & ERNIE Bot & ChatGPT & Bard   \\
\midrule
\multirow{3}{*}{Completeness}        & Zero-shot          & \textbf{5.00} & \textbf{5.00} & 4.99 & 4.99\\
                                     & One-shot           & 4.99 & \textbf{5.00} & \textbf{5.00} & 4.98\\
                                     & Three-shot         & 4.96 & \textbf{5.00} & \textbf{5.00} & 4.97\\
\cmidrule{2-6}
\multirow{3}{*}{Correctness}         & Zero-shot          & 4.40 & 4.08 & 4.13 & 4.15\\
                                     & One-shot           & 4.45 & 4.50 & 4.32 & 4.29\\
                                     & Three-shot         & 4.37 & 4.47 & 4.66 & \textbf{4.74}\\
\cmidrule{2-6}
\multirow{3}{*}{Conciseness}         & Zero-shot          & 2.00 & 3.13 & 3.10 & 3.22\\
                                     & One-shot           & 4.08 & 4.17 & 4.14 & 4.24\\
                                     & Three-shot         & 4.37 & 4.39 & \textbf{4.62} & 4.36\\
\cmidrule{2-6}
\multirow{3}{*}{Verisimilitude}      & Zero-shot          & 3.00 & 4.01 & 4.06 & 4.06\\
                                     & One-shot           & 4.31 & 4.37 & 4.40 & 4.41\\
                                     & Three-shot         & 4.42 & 4.45 & 4.64 & \textbf{4.65}\\
\cmidrule{2-6}
\multirow{3}{*}{Replaceability}      & Zero-shot          & 3.83 & 4.01 & 4.05 & 4.04\\
                                     & One-shot           & 4.11 & 4.33 & 4.22 & 4.16\\
                                     & Three-shot         & 4.26 & 4.45 & \textbf{4.56} & 4.47\\
\botrule
\end{tabular*}
\end{table}

\begin{figure}[h]
\centering
\includegraphics[width=1.0\textwidth]{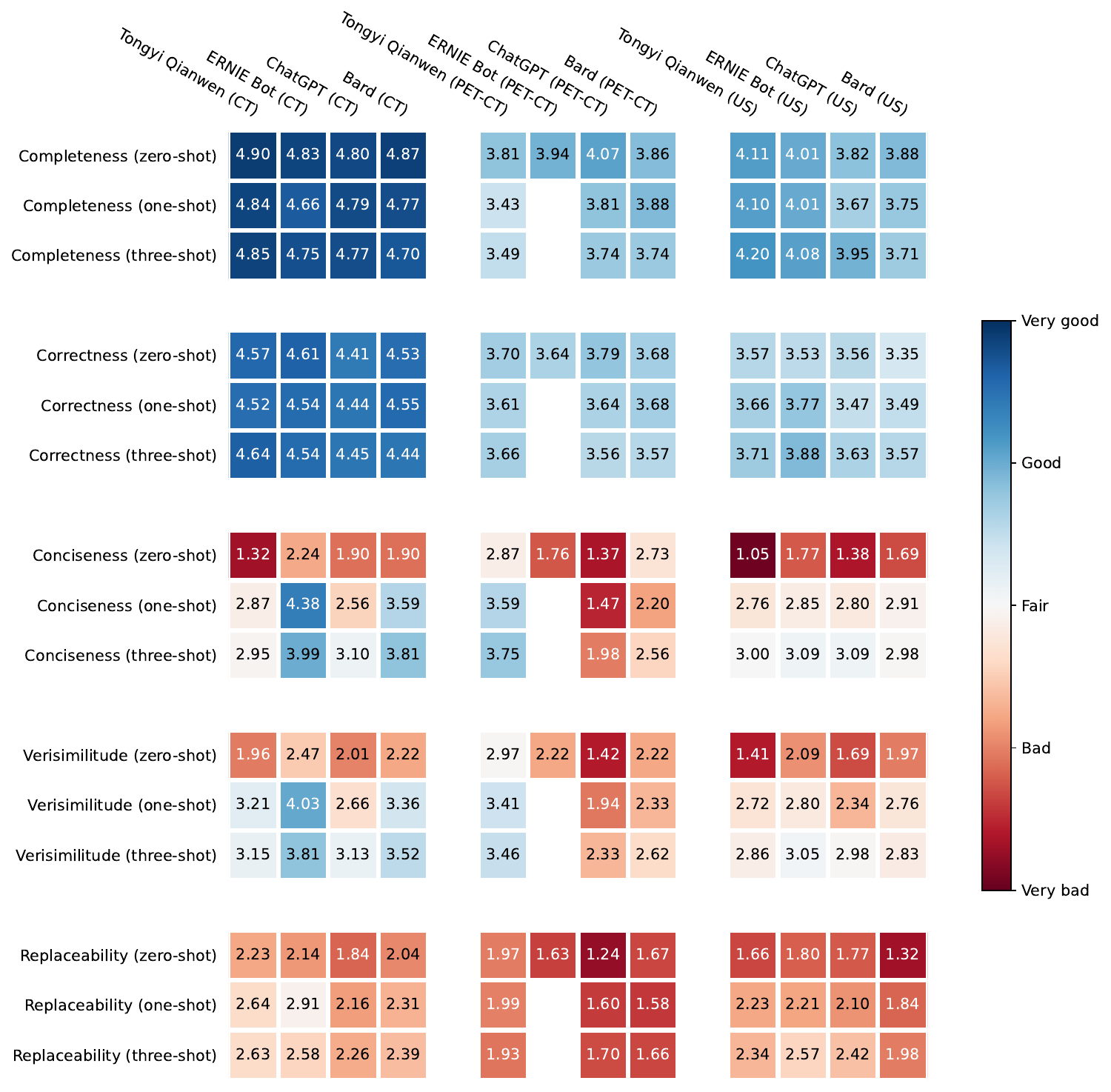}
\caption{Clinician I's evaluation results.}\label{figA1}
\end{figure}

\begin{figure}[h]
\centering
\includegraphics[width=1.0\textwidth]{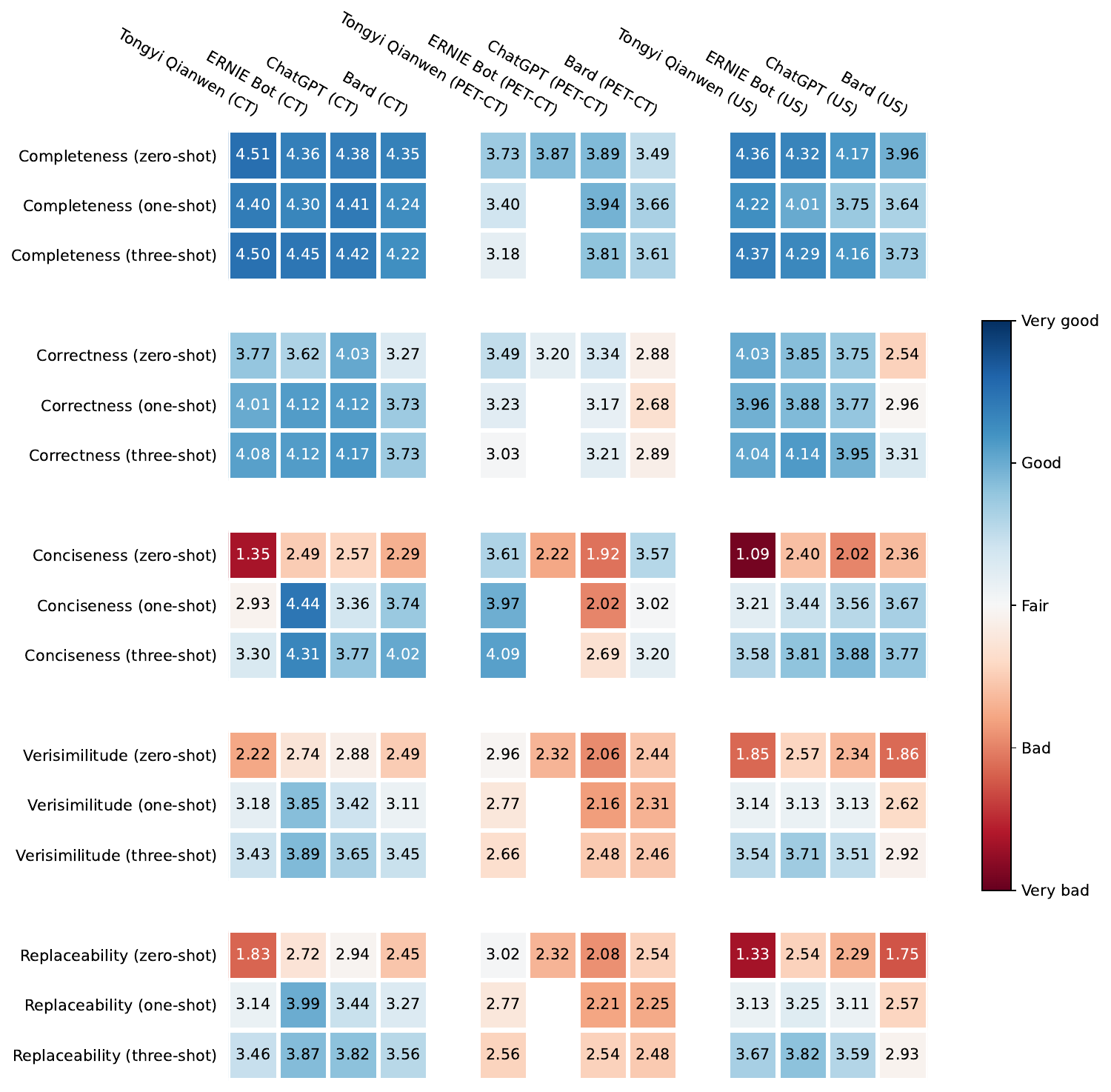}
\caption{Clinician II's evaluation results.}\label{figA2}
\end{figure}

\begin{figure}[h]
\centering
\includegraphics[width=1.0\textwidth]{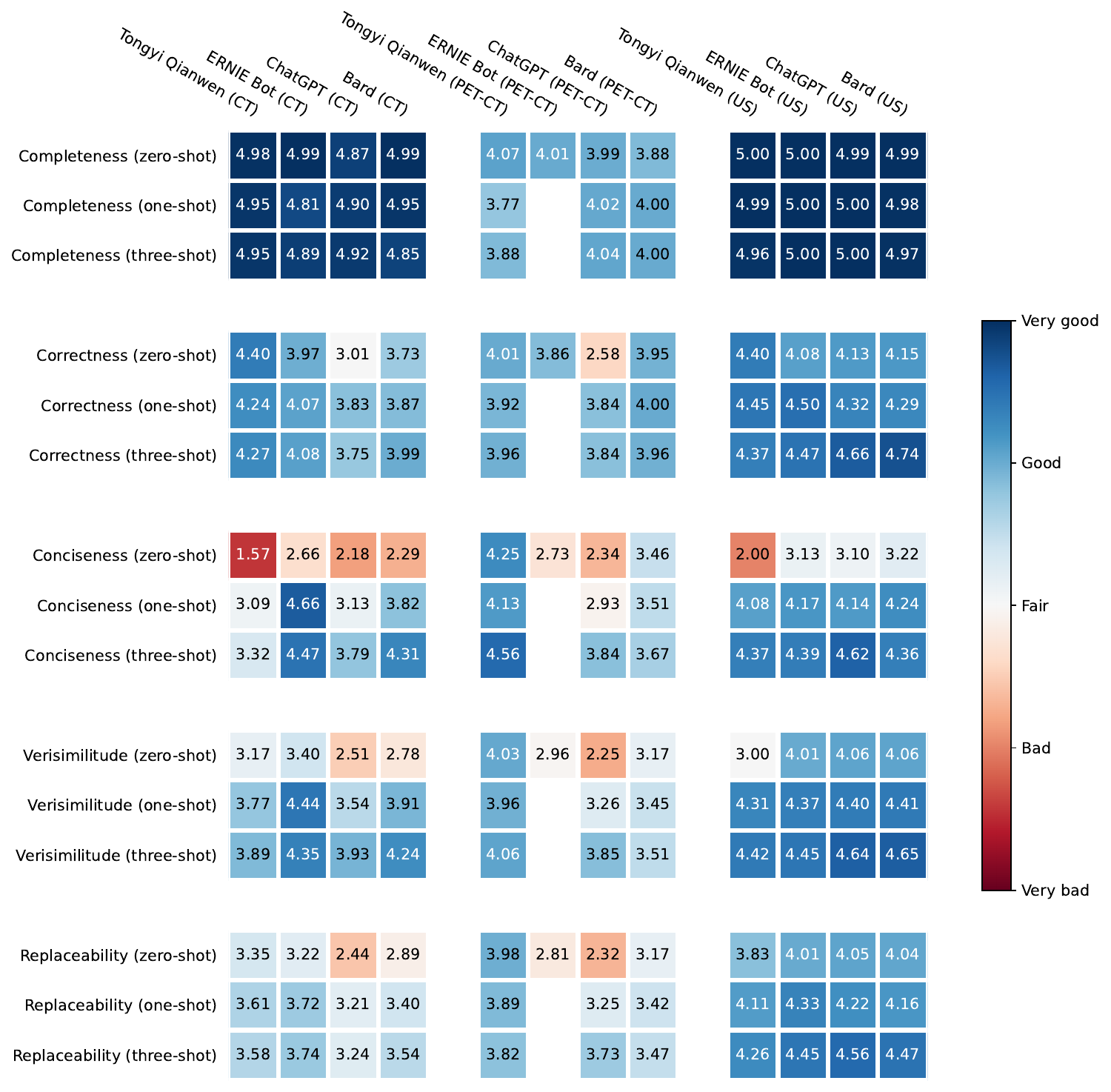}
\caption{Clinician III's evaluation results.}\label{figA3}
\end{figure}

\begin{figure}[h]
\centering
\includegraphics[width=1.0\textwidth]{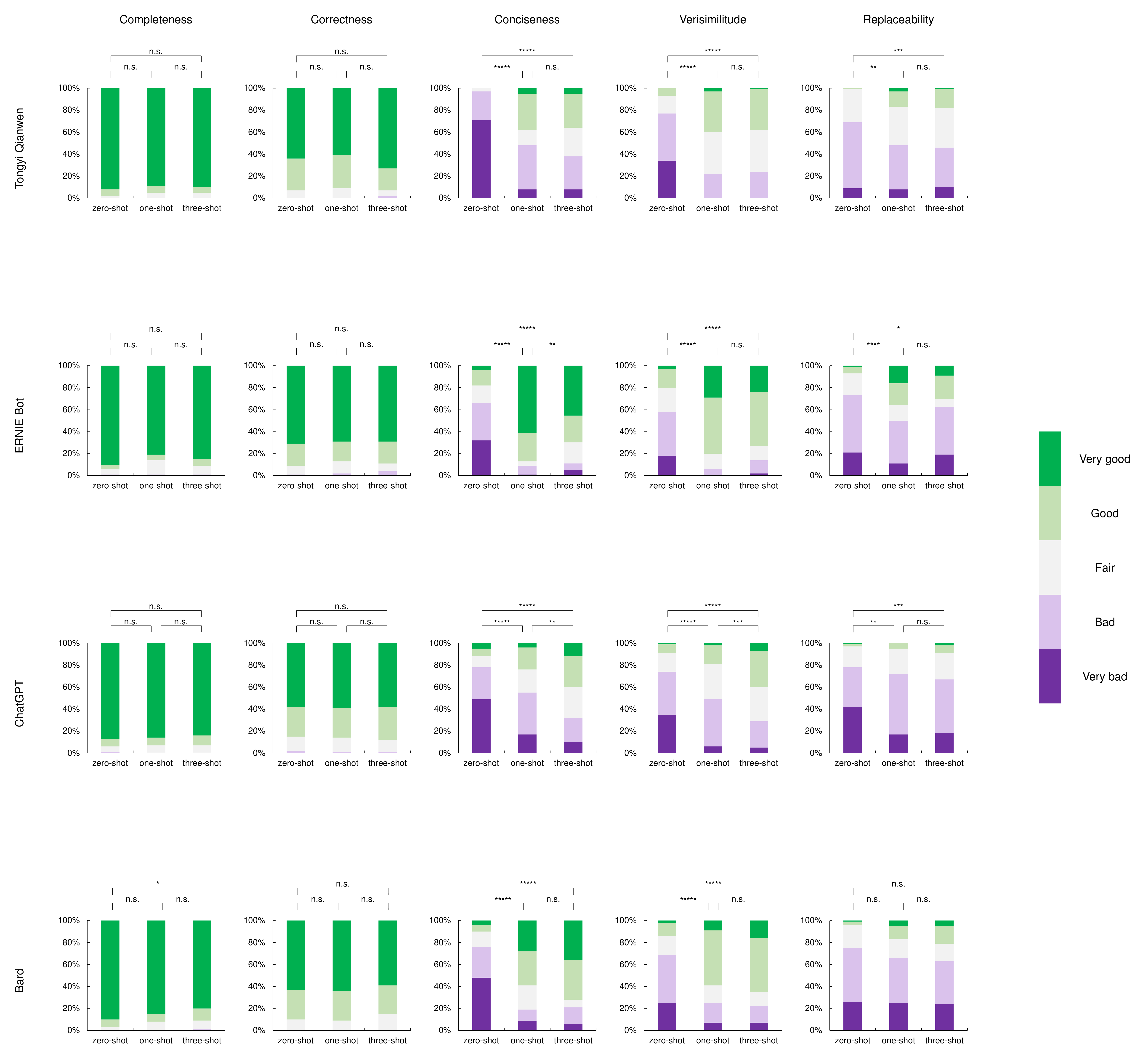}
\caption{Statistical test of Clinician I's CT evaluation.}\label{figA4}
\end{figure}

\begin{figure}[h]
\centering
\includegraphics[width=1.0\textwidth]{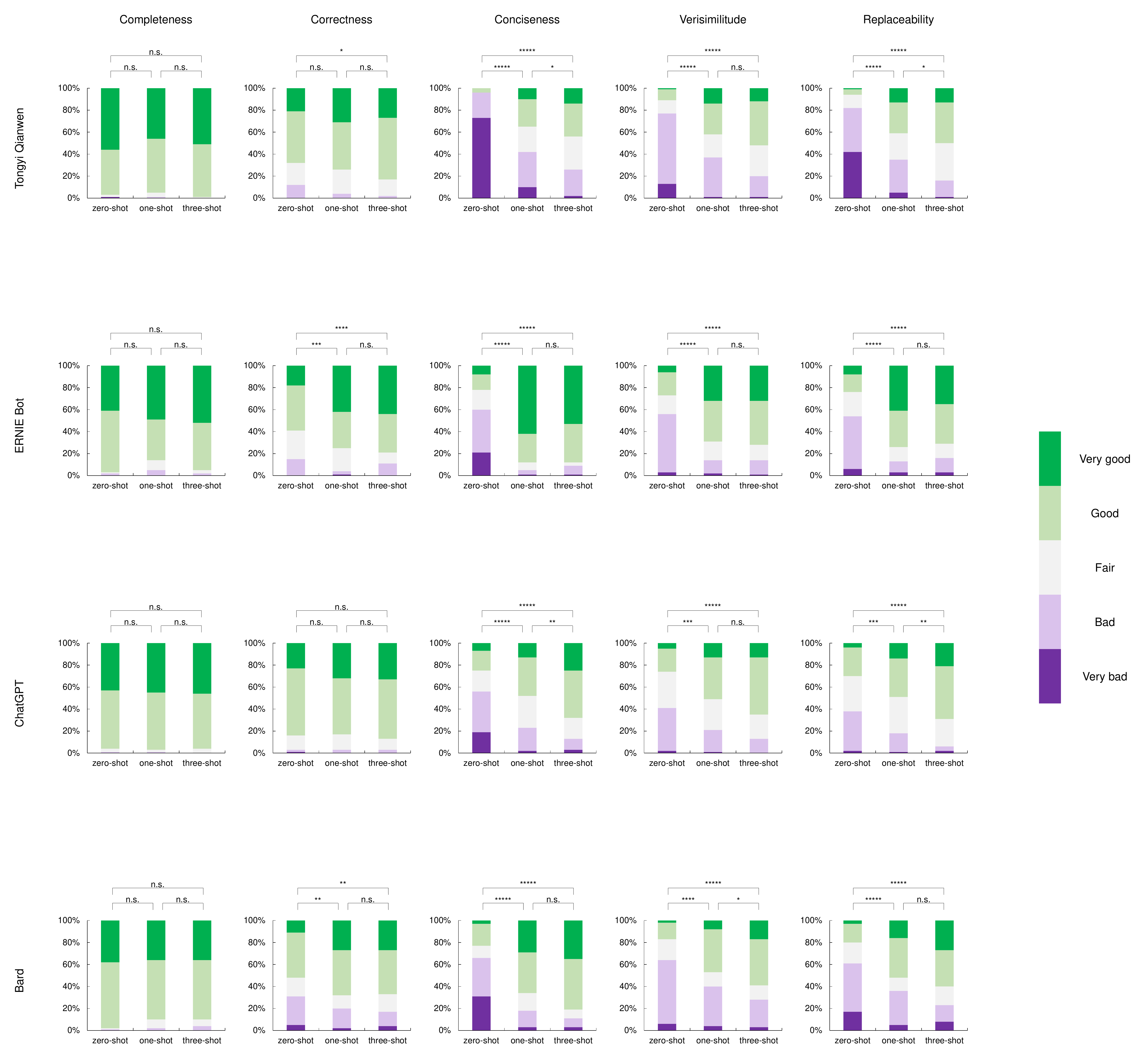}
\caption{Statistical test of Clinician II's CT evaluation.}\label{figA5}
\end{figure}

\begin{figure}[h]
\centering
\includegraphics[width=1.0\textwidth]{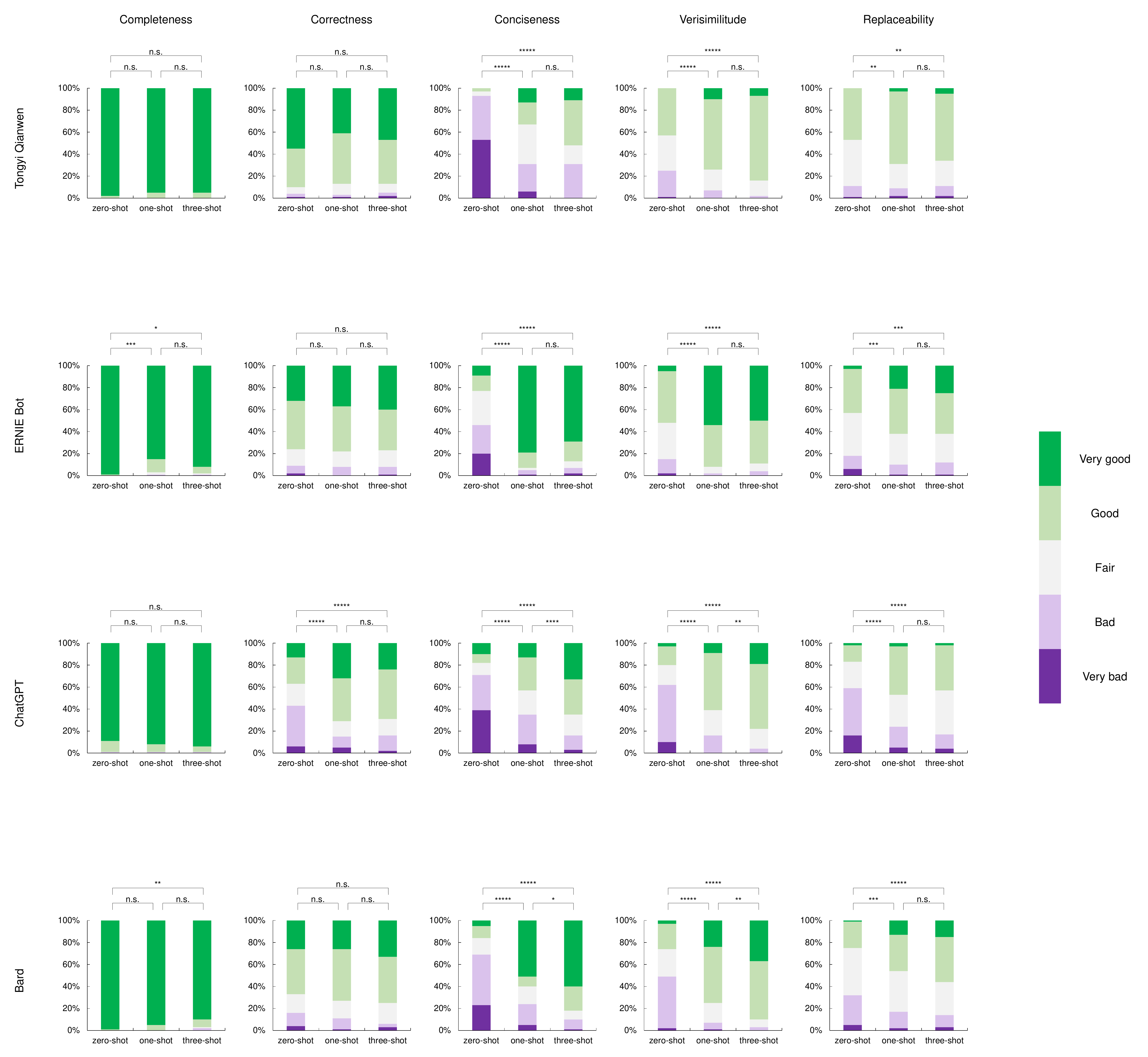}
\caption{Statistical test of Clinician III's CT evaluation.}\label{figA6}
\end{figure}

\begin{figure}[h]
\centering
\includegraphics[width=1.0\textwidth]{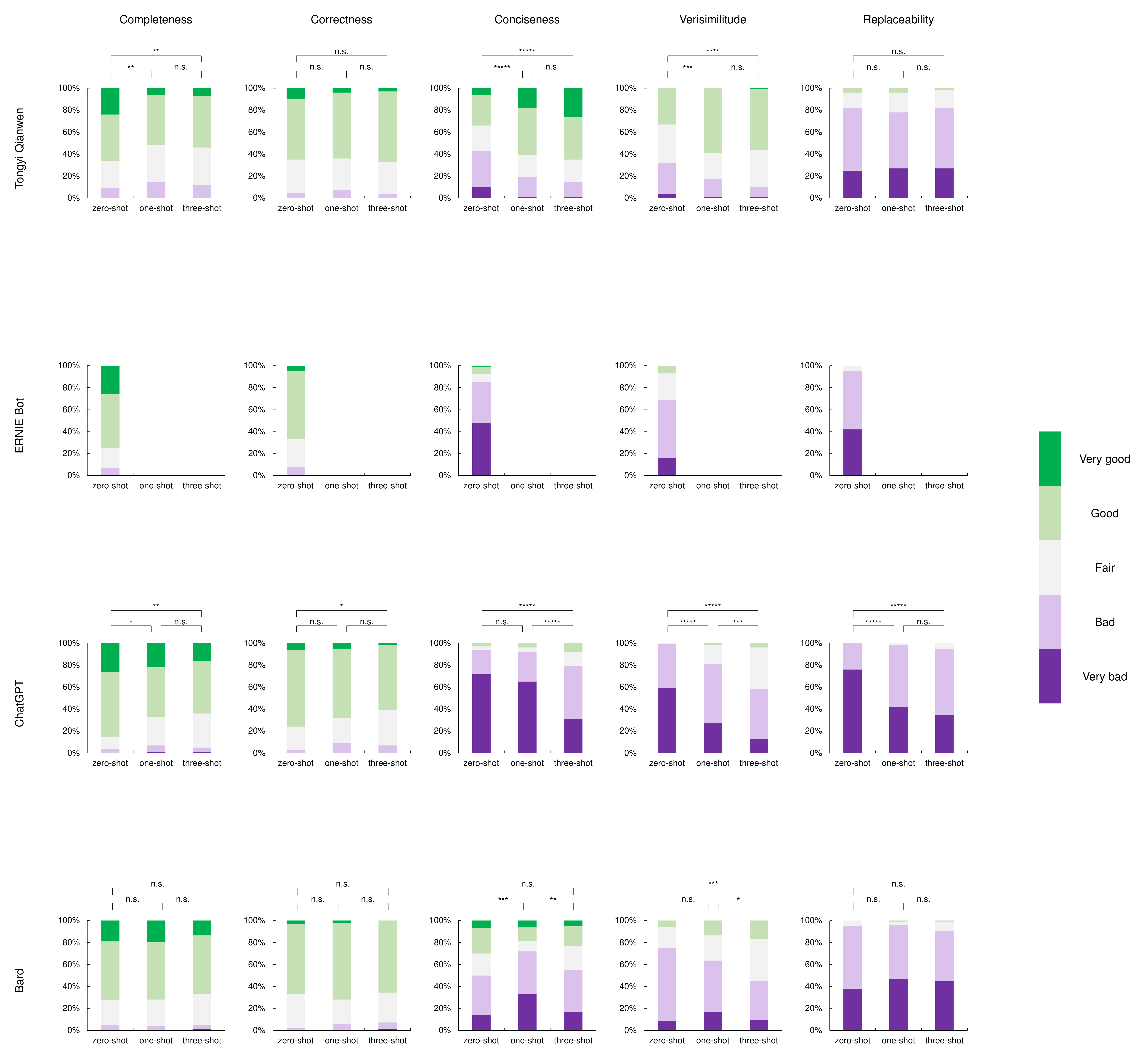}
\caption{Statistical test of Clinician I's PETCT evaluation.}\label{figA7}
\end{figure}

\begin{figure}[h]
\centering
\includegraphics[width=1.0\textwidth]{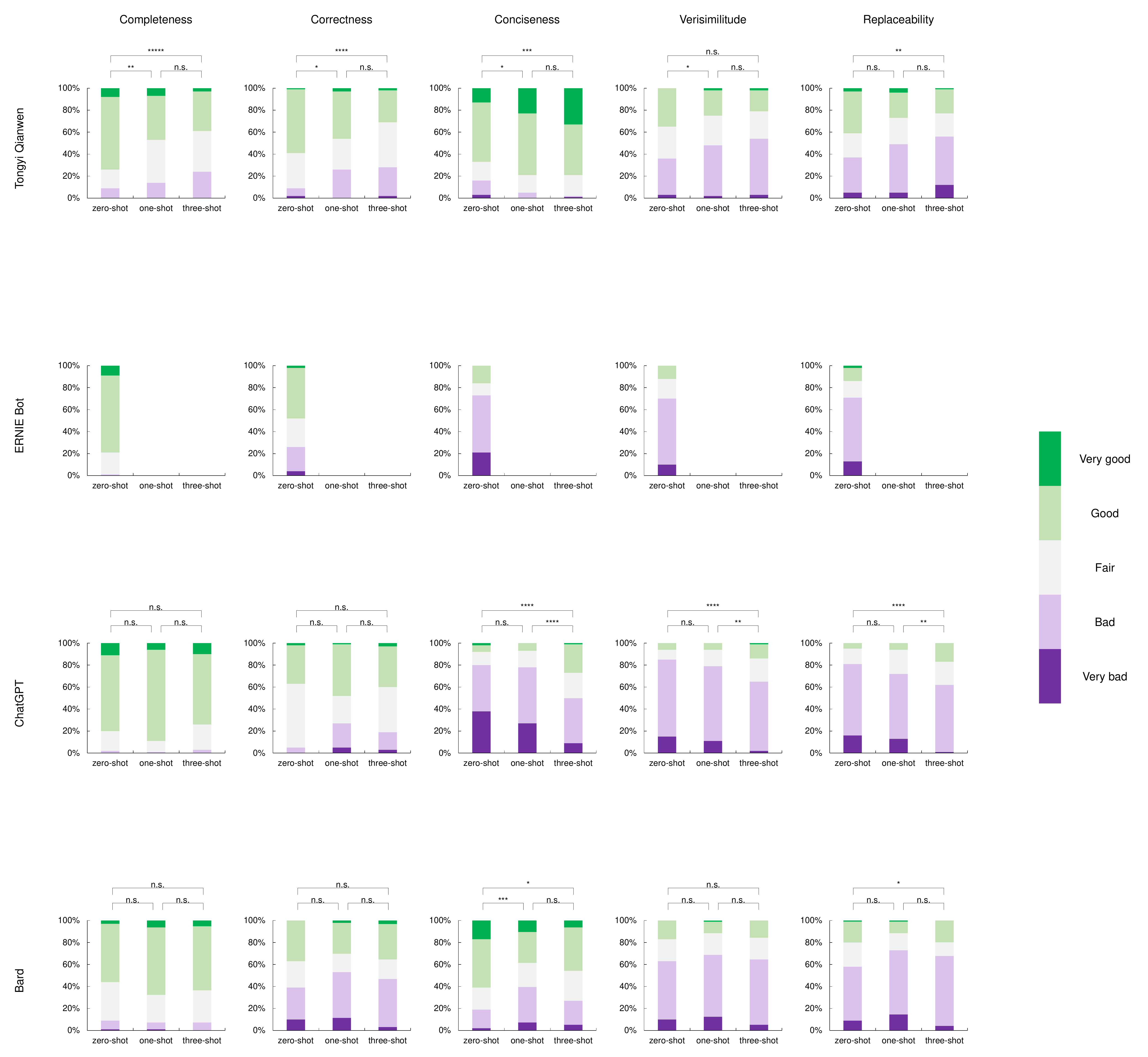}
\caption{Statistical test of Clinician II's PETCT evaluation.}\label{figA8}
\end{figure}

\begin{figure}[h]
\centering
\includegraphics[width=1.0\textwidth]{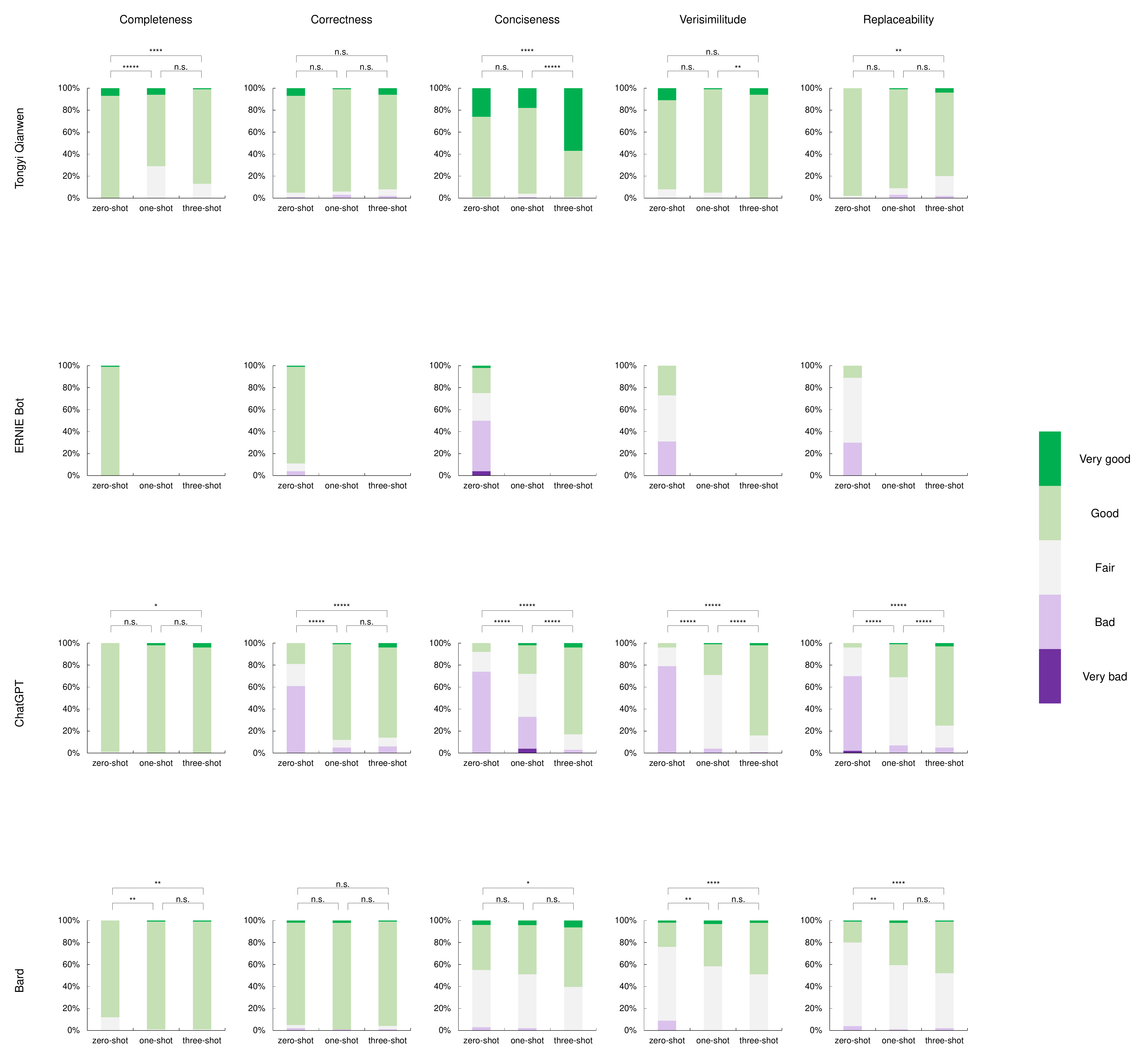}
\caption{Statistical test of Clinician III's PETCT evaluation.}\label{figA9}
\end{figure}

\begin{figure}[h]
\centering
\includegraphics[width=1.0\textwidth]{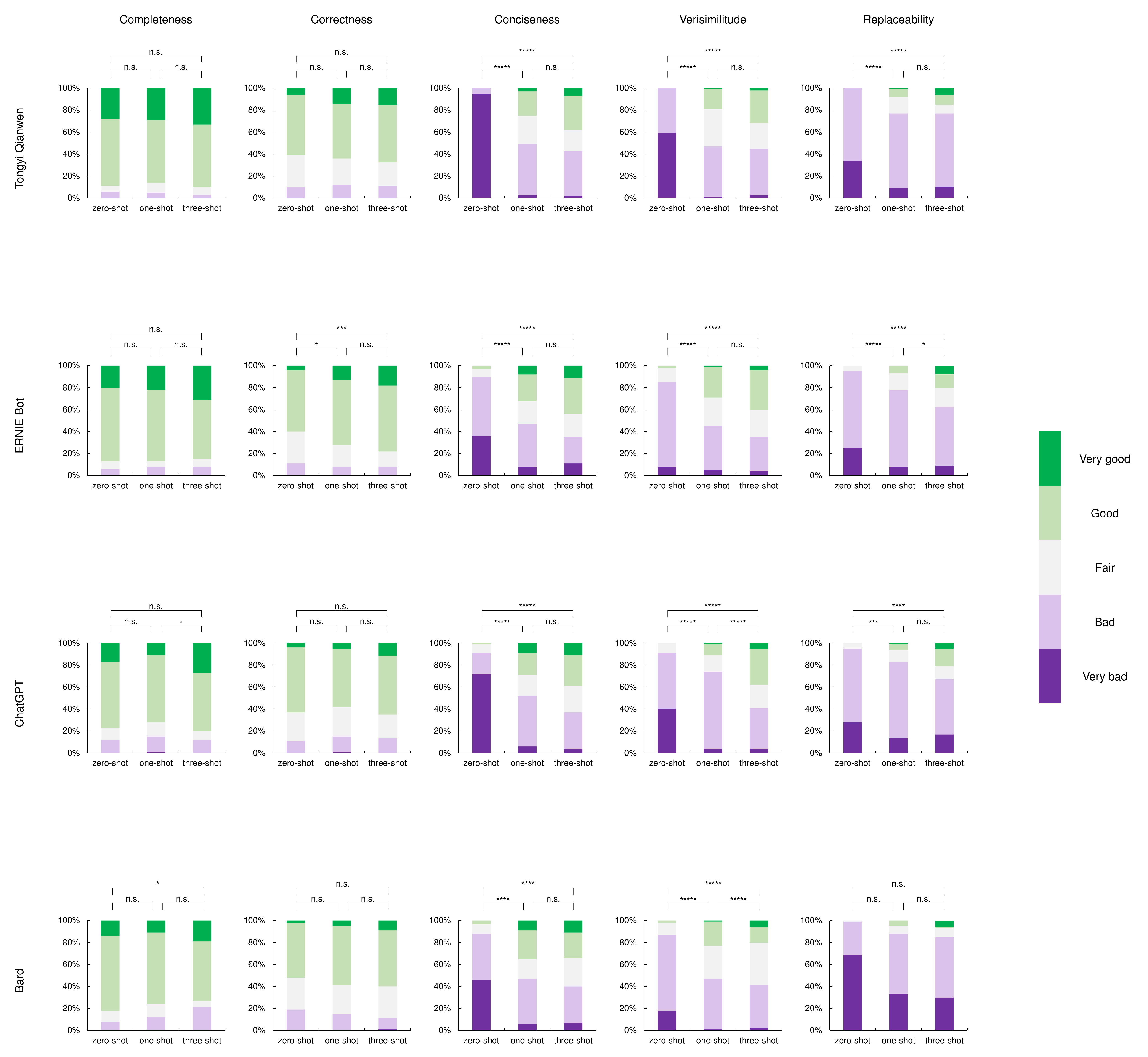}
\caption{Statistical test of Clinician I's US evaluation.}\label{figA10}
\end{figure}

\begin{figure}[h]
\centering
\includegraphics[width=1.0\textwidth]{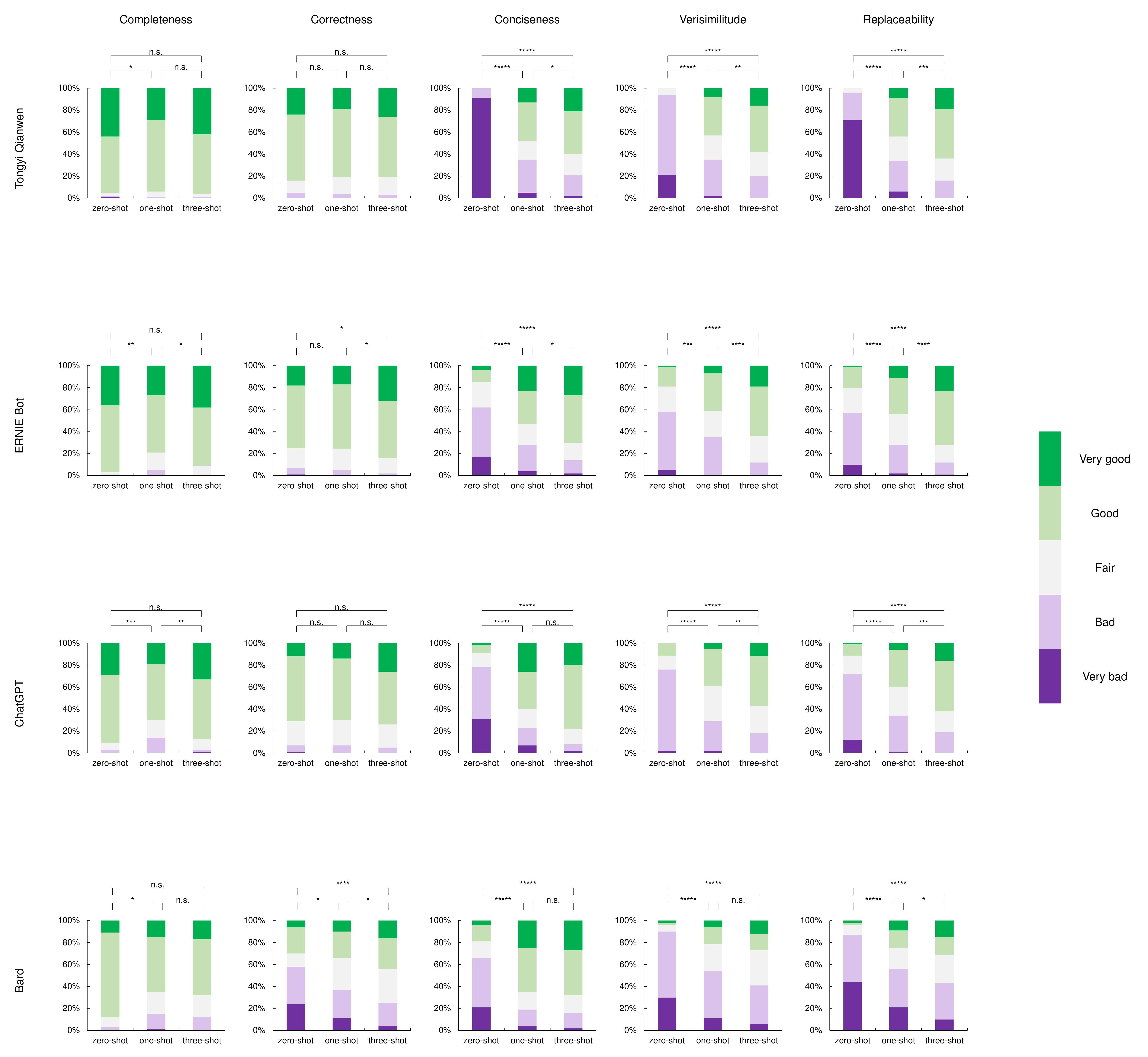}
\caption{Statistical test of Clinician II's US evaluation.}\label{figA11}
\end{figure}

\begin{figure}[h]
\centering
\includegraphics[width=1.0\textwidth]{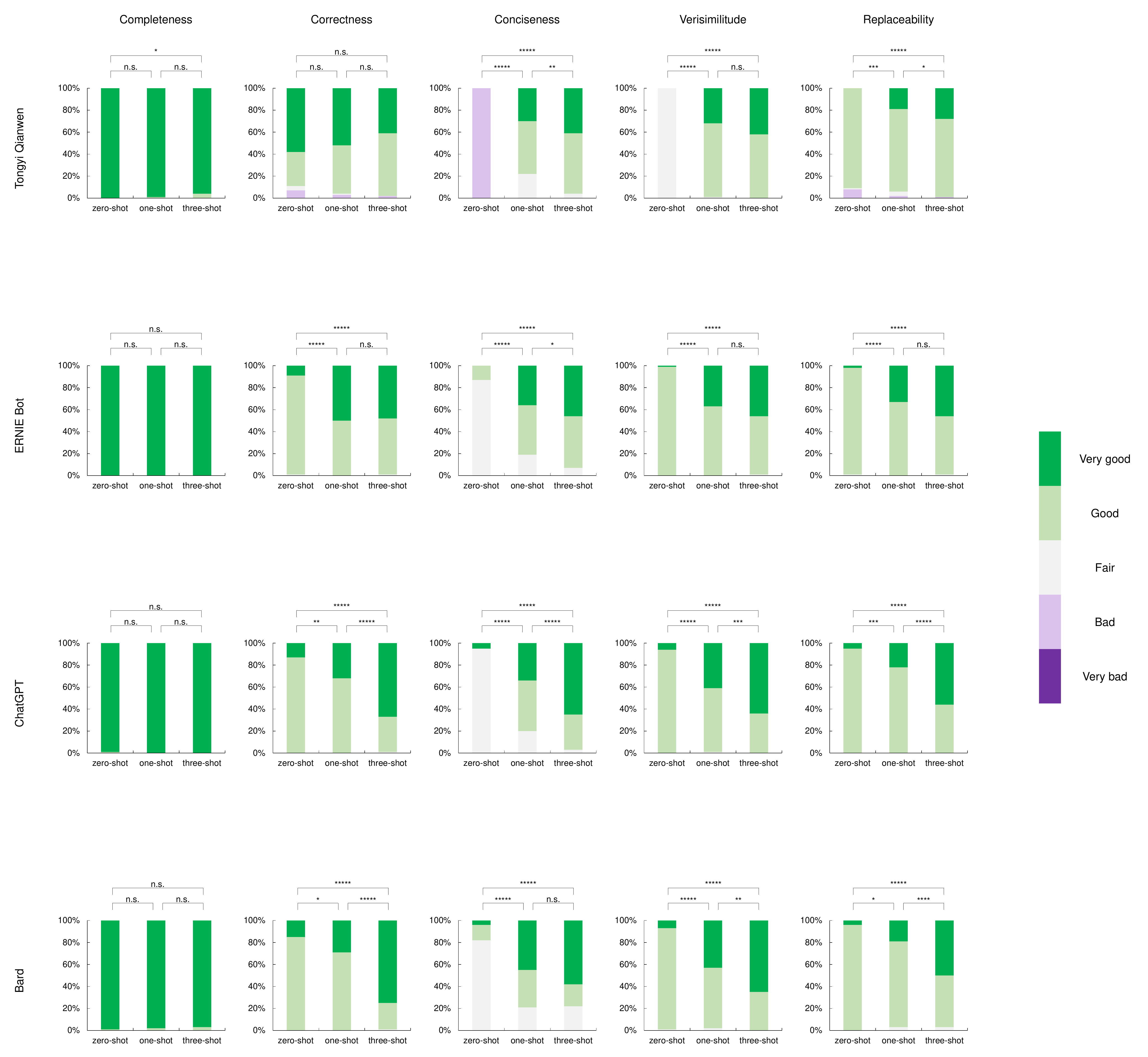}
\caption{Statistical test of Clinician III's US evaluation.}\label{figA12}
\end{figure}

\begin{table}[h]
\tiny
\caption{Automatic quantitative evaluation results of the generated raw CT impressions.}\label{tableA10}
\begin{tabular*}{\textwidth}{@{\extracolsep\fill}llcccccc}
\toprule%
Prompt type                          & Model              & BLEU1 & BLEU2 & BLEU3 & BLEU4 & ROUGE-L & METEOR  \\
\midrule
\multirow{8}{*}{Zero-shot}           & Tongyi Qianwen     & 0.017 & 0.014 & 0.011 & 0.009 & 0.205 & 0.147\\
                                     & ERNIE Bot          & 0.089 & 0.073 & 0.061 & 0.052 & 0.291 & 0.234\\
                                     & ChatGPT            & 0.028 & 0.021 & 0.017 & 0.014 & 0.203 & 0.150\\
                                     & Bard               & 0.000 & 0.000 & 0.000 & 0.000 & 0.112 & 0.070\\
                                     & Baichuan           & 0.030 & 0.023 & 0.018 & 0.015 & 0.195 & 0.138\\
                                     & ChatGLM            & 0.009 & 0.007 & 0.005 & 0.004 & 0.174 & 0.125\\
                                     & HuatuoGPT          & 0.009 & 0.007 & 0.005 & 0.004 & 0.169 & 0.128\\
                                     & ChatGLM-Med        & 0.171 & 0.116 & 0.084 & 0.064 & 0.154 & 0.156\\
\cmidrule{2-8}
\multirow{8}{*}{One-shot}            & Tongyi Qianwen     & 0.171 & 0.139 & 0.115 & 0.096 & 0.320 & 0.259\\
                                     & \textbf{ERNIE Bot}          & \textbf{0.431} & \textbf{0.356} & \textbf{0.302} & \textbf{0.257} & \textbf{0.463} & \textbf{0.458}\\
                                     & ChatGPT            & 0.118 & 0.094 & 0.077 & 0.063 & 0.292 & 0.244\\
                                     & Bard               & 0.001 & 0.000 & 0.000 & 0.000 & 0.133 & 0.086\\
                                     & Baichuan           & 0.014 & 0.011 & 0.009 & 0.007 & 0.181 & 0.135\\
                                     & ChatGLM            & 0.042 & 0.033 & 0.027 & 0.022 & 0.223 & 0.181\\
                                     & HuatuoGPT          & 0.004 & 0.003 & 0.002 & 0.002 & 0.195 & 0.157\\
                                     & ChatGLM-Med        & 0.189 & 0.132 & 0.098 & 0.076 & 0.167 & 0.166\\
\cmidrule{2-8}
\multirow{8}{*}{Three-shot}          & Tongyi Qianwen     & 0.224 & 0.183 & 0.153 & 0.129 & 0.350 & 0.299\\
                                     & ERNIE Bot          & 0.310 & 0.257 & 0.217 & 0.184 & 0.429 & 0.402\\
                                     & ChatGPT            & 0.166 & 0.136 & 0.114 & 0.096 & 0.337 & 0.298\\
                                     & Bard               & 0.001 & 0.001 & 0.000 & 0.000 & 0.137 & 0.090\\
                                     & Baichuan           & 0.115 & 0.091 & 0.075 & 0.062 & 0.293 & 0.256\\
                                     & ChatGLM            & 0.045 & 0.035 & 0.029 & 0.023 & 0.225 & 0.178\\
                                     & HuatuoGPT          & 0.000 & 0.000 & 0.000 & 0.000 & 0.148 & 0.108\\
                                     & ChatGLM-Med        & 0.150 & 0.095 & 0.063 & 0.045 & 0.140 & 0.140\\
\botrule
\end{tabular*}
\end{table}

\begin{table}[h]
\tiny
\caption{Automatic quantitative evaluation results of the generated raw PET-CT impressions.}\label{tableA11}
\begin{tabular*}{\textwidth}{@{\extracolsep\fill}llcccccc}
\toprule%
Prompt type                          & Model              & BLEU1 & BLEU2 & BLEU3 & BLEU4 & ROUGE-L & METEOR  \\
\midrule
\multirow{8}{*}{Zero-shot}           & Tongyi Qianwen     & 0.438 & 0.335 & 0.264 & 0.212 & 0.316 & 0.382\\
                                     & ERNIE Bot          & 0.318 & 0.247 & 0.201 & 0.167 & 0.302 & 0.326\\
                                     & ChatGPT            & 0.159 & 0.119 & 0.093 & 0.074 & 0.226 & 0.227\\
                                     & Bard               & 0.113 & 0.082 & 0.064 & 0.051 & 0.210 & 0.197\\
                                     & Baichuan           & 0.100 & 0.076 & 0.060 & 0.048 & 0.216 & 0.212\\
                                     & ChatGLM            & 0.103 & 0.077 & 0.061 & 0.049 & 0.208 & 0.206\\
                                     & HuatuoGPT          & 0.110 & 0.082 & 0.066 & 0.054 & 0.203 & 0.193\\
                                     & ChatGLM-Med        & 0.090 & 0.059 & 0.044 & 0.035 & 0.094 & 0.145\\
\cmidrule{2-8}
\multirow{8}{*}{One-shot}            & \textbf{Tongyi Qianwen}     & \textbf{0.476} & \textbf{0.368} & \textbf{0.294} & \textbf{0.240} &                                      0.340 & 0.424\\
                                     & ERNIE Bot          & - & - & - & - & - & -\\
                                     & ChatGPT            & 0.221 & 0.166 & 0.129 & 0.102 & 0.248 & 0.266\\
                                     & Bard               & 0.063 & 0.048 & 0.038 & 0.031 & 0.196 & 0.174\\
                                     & Baichuan           & 0.066 & 0.051 & 0.041 & 0.034 & 0.210 & 0.197\\
                                     & ChatGLM            & 0.199 & 0.149 & 0.117 & 0.094 & 0.232 & 0.246\\
                                     & HuatuoGPT          & 0.083 & 0.062 & 0.048 & 0.039 & 0.188 & 0.172\\
                                     & ChatGLM-Med        & - & - & - & - & - & -\\
\cmidrule{2-8}
\multirow{8}{*}{Three-shot}          & \textbf{Tongyi Qianwen}     & 0.439 & 0.340 & 0.272 & 0.221 & \textbf{0.354} & \textbf{0.450}\\
                                     & ERNIE Bot          & - & - & - & - & - & -\\
                                     & ChatGPT            & 0.309 & 0.234 & 0.183 & 0.147 & 0.274 & 0.309\\
                                     & Bard               & 0.105 & 0.080 & 0.064 & 0.052 & 0.217 & 0.204\\
                                     & Baichuan           & - & - & - & - & - & -\\
                                     & ChatGLM            & 0.188 & 0.143 & 0.115 & 0.094 & 0.215 & 0.227\\
                                     & HuatuoGPT          & - & - & - & - & - & -\\
                                     & ChatGLM-Med        & - & - & - & - & - & -\\
\botrule
\end{tabular*}
\end{table}

\begin{table}[h]
\tiny
\caption{Automatic quantitative evaluation results of the generated raw Ultrasound impressions.}\label{tableA12}
\begin{tabular*}{\textwidth}{@{\extracolsep\fill}llcccccc}
\toprule%
Prompt type                          & Model              & BLEU1 & BLEU2 & BLEU3 & BLEU4 & ROUGE-L & METEOR  \\
\midrule
\multirow{8}{*}{Zero-shot}           & Tongyi Qianwen     & 0.000 & 0.000 & 0.000 & 0.000 & 0.122 & 0.083\\
                                     & ERNIE Bot          & 0.014 & 0.013 & 0.011 & 0.010 & 0.252 & 0.187\\
                                     & ChatGPT            & 0.001 & 0.001 & 0.001 & 0.001 & 0.157 & 0.105\\
                                     & Bard               & 0.000 & 0.000 & 0.000 & 0.000 & 0.090 & 0.053\\
                                     & Baichuan           & 0.006 & 0.006 & 0.005 & 0.004 & 0.231 & 0.167\\
                                     & ChatGLM            & 0.001 & 0.001 & 0.001 & 0.000 & 0.163 & 0.109\\
                                     & HuatuoGPT          & 0.000 & 0.000 & 0.000 & 0.000 & 0.118 & 0.073\\
                                     & ChatGLM-Med        & 0.165 & 0.131 & 0.109 & 0.093 & 0.162 & 0.138\\
\cmidrule{2-8}
\multirow{8}{*}{One-shot}            & Tongyi Qianwen     & 0.122 & 0.101 & 0.082 & 0.066 & 0.319 & 0.294\\
                                     & ERNIE Bot          & 0.072 & 0.062 & 0.055 & 0.048 & 0.304 & 0.257\\
                                     & ChatGPT            & 0.008 & 0.007 & 0.006 & 0.006 & 0.239 & 0.193\\
                                     & Bard               & 0.000 & 0.000 & 0.000 & 0.000 & 0.106 & 0.065\\
                                     & Baichuan           & 0.049 & 0.042 & 0.036 & 0.032 & 0.278 & 0.226\\
                                     & ChatGLM            & 0.005 & 0.004 & 0.003 & 0.003 & 0.190 & 0.137\\
                                     & HuatuoGPT          & 0.000 & 0.000 & 0.000 & 0.000 & 0.104 & 0.073\\
                                     & \textbf{ChatGLM-Med}        & \textbf{0.165} & 0.131 & 0.111 & 0.097 & 0.143 & 0.124\\
\cmidrule{2-8}
\multirow{8}{*}{Three-shot}          & Tongyi Qianwen     & 0.150 & 0.131 & 0.115 & 0.100 & 0.408 & 0.365\\
                                     & ERNIE Bot          & 0.042 & 0.038 & 0.035 & 0.033 & 0.351 & 0.286\\
                                     & \textbf{ChatGPT}            & 0.151 & \textbf{0.137} & \textbf{0.126} & \textbf{0.115} & \textbf{0.441} & \textbf{0.396}\\
                                     & Bard               & 0.000 & 0.000 & 0.000 & 0.000 & 0.111 & 0.068\\
                                     & Baichuan           & 0.098 & 0.081 & 0.069 & 0.059 & 0.270 & 0.224\\
                                     & ChatGLM            & 0.003 & 0.002 & 0.002 & 0.002 & 0.167 & 0.115\\
                                     & HuatuoGPT          & 0.000 & 0.000 & 0.000 & 0.000 & 0.096 & 0.062\\
                                     & ChatGLM-Med        & 0.131 & 0.107 & 0.093 & 0.083 & 0.141 & 0.118\\
\botrule
\end{tabular*}
\end{table}

\begin{figure}[h]
\centering
\includegraphics[width=1\textwidth]{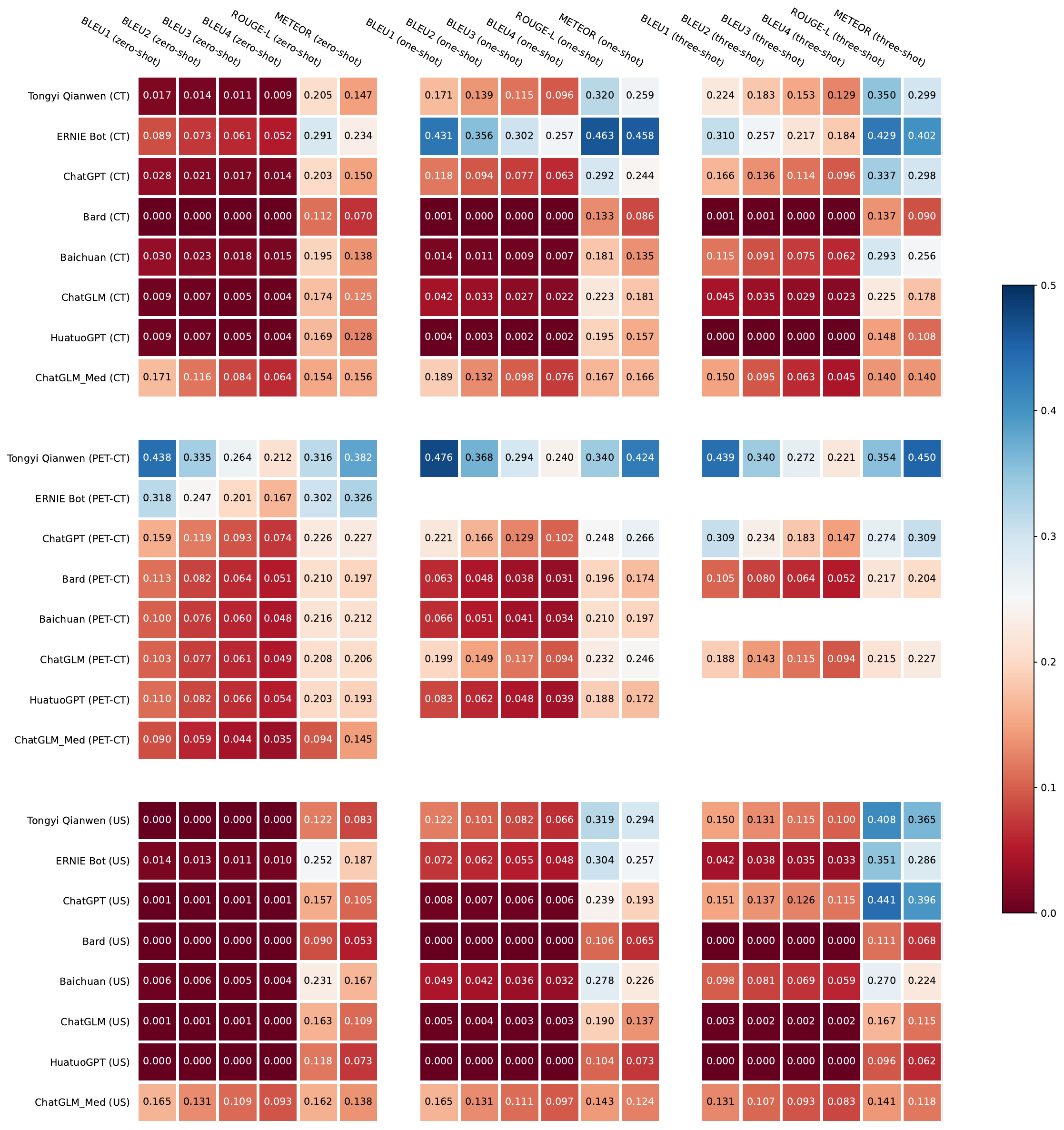}
\caption{Automatic quantitative evaluation results of the generated raw impressions.}\label{figA13}
\end{figure}

\begin{landscape}

\begin{figure}[h]
\centering
\includegraphics[width=1.5\textwidth]{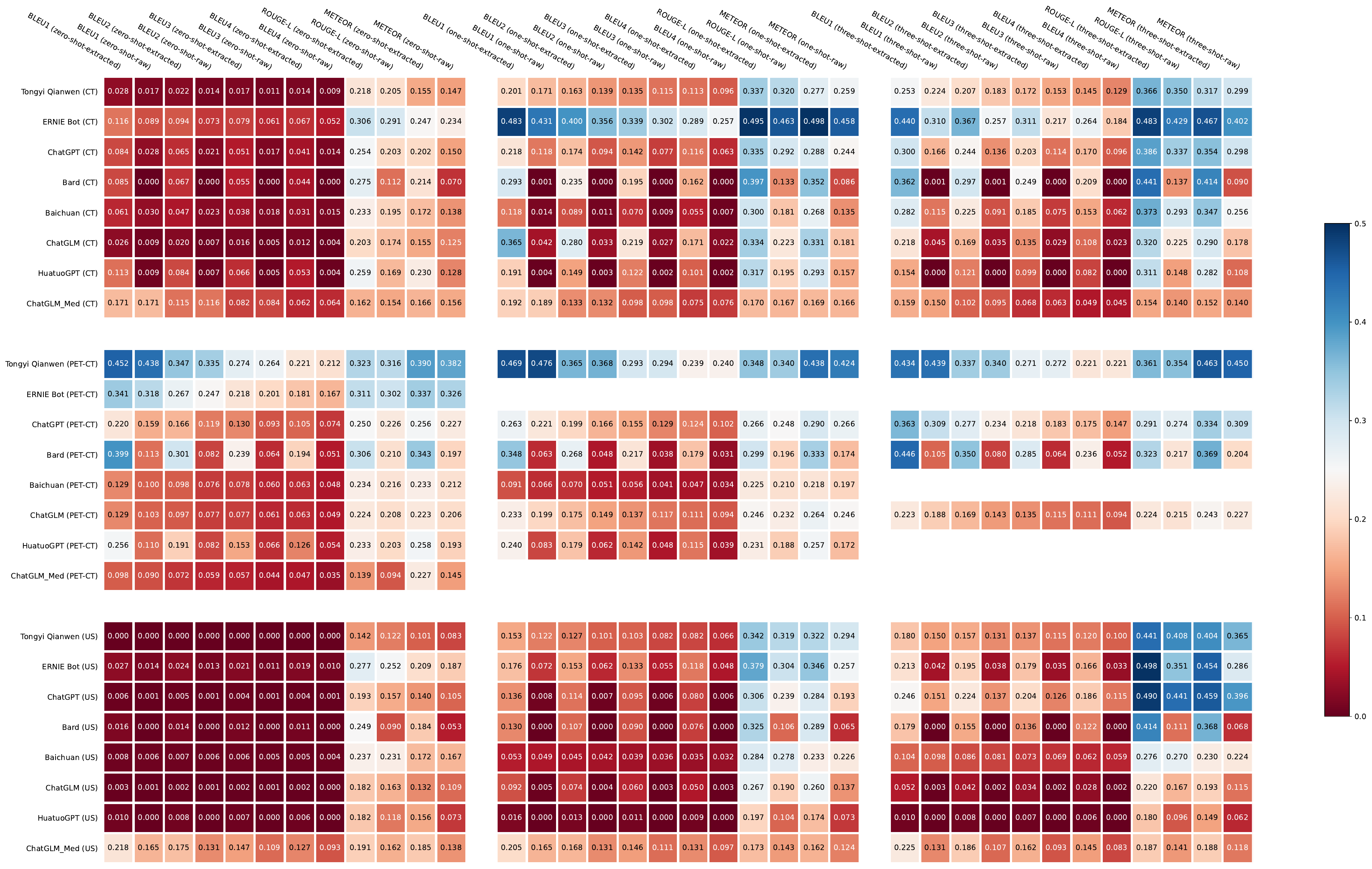}
\caption{Heatmap of the automatic quantitative evaluation results obtained by evaluating extracted impressions and raw outputs.}\label{figA14}
\end{figure}

\end{landscape}

\begin{figure}[h]
\centering
\includegraphics[width=1\textwidth]{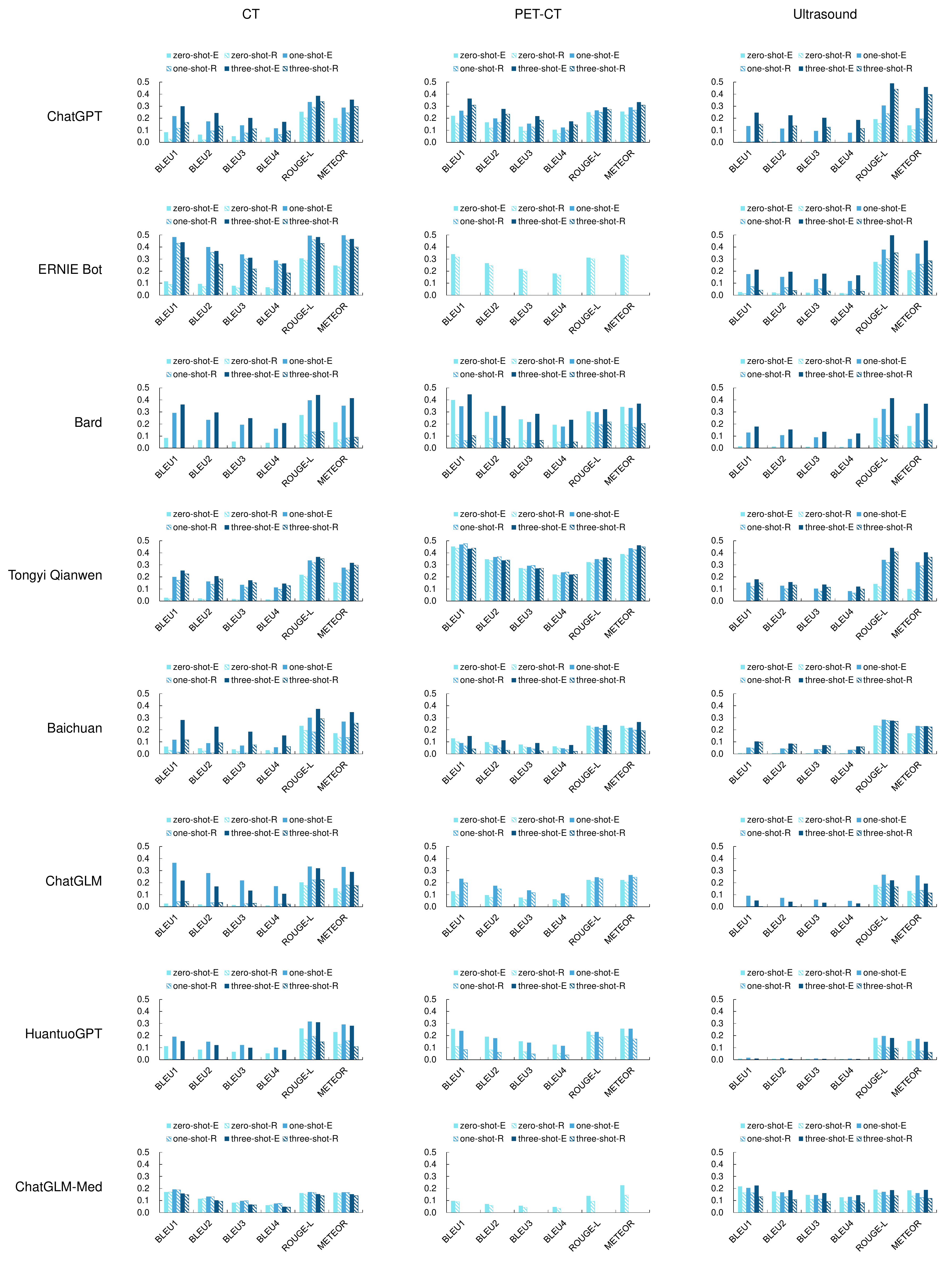}
\caption{Bar chart of the automatic quantitative evaluation results obtained by evaluating extracted impressions and raw outputs.}\label{figA15}
\end{figure}

\end{appendices}


% \bibliography{sn-bibliography}% common bib file
% %% if required, the content of .bbl file can be included here once bbl is generated
% %%\input sn-article.bbl

\end{document}